\documentclass[lettersize,journal]{IEEEtran}
\usepackage{amsmath,amsfonts}
\usepackage{algorithmic}
\usepackage{algorithm}
\usepackage{bm}

\usepackage{amssymb}
\usepackage{subcaption}

\usepackage{booktabs}
\usepackage{multirow}
\usepackage{adjustbox}
\usepackage{tablefootnote}
\usepackage{threeparttable}
\usepackage{bbm}
\usepackage{svg}

\usepackage{mathtools}
\usepackage{array}
\usepackage{textcomp}
\usepackage{stfloats}
\usepackage{url}
\usepackage{verbatim}
\usepackage{graphicx}
\usepackage{cite}
\usepackage{caption} 

\usepackage[colorlinks=true, linkcolor=blue, citecolor=blue]{hyperref}
\let\oldtwocolumn\twocolumn
\renewcommand\twocolumn[1][]{%
    \oldtwocolumn[{#1}{
    \begin{center}
           \includegraphics[width=\textwidth]{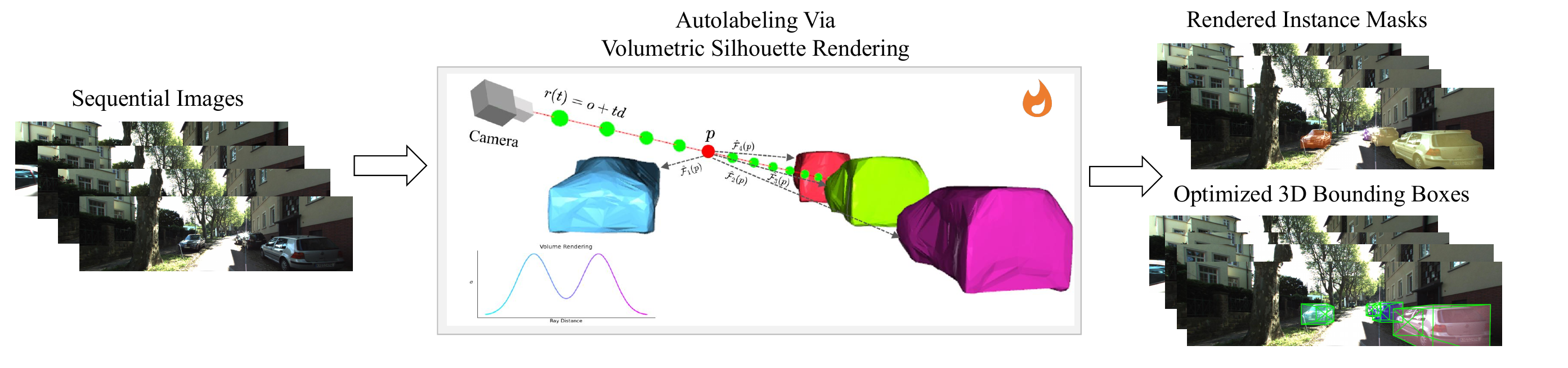}
           \captionof{figure}{Autolabeling via the proposed volumetric silhouette rendering, without relying on 3D annotations or ground truth LiDAR point clouds.}
           \label{fig:teaser}
        \end{center}
    }]
}

\begin{document}

% change title

\title{ VSRD++: Autolabeling for 3D Object Detection via Instance-Aware Volumetric Silhouette Rendering}

\author{ Zihua Liu,~\IEEEmembership{Student Member,~IEEE} , Hiroki Sakuma~\IEEEmembership{Member,~IEEE},  and Masatoshi Okutomi,~\IEEEmembership{Member,~IEEE}
        % <-this % stops a space
\thanks{Z.Liu, H.Sakuma and M.Okutomi are with the Department of System and Control Engineering, Institute of Science Tokyo, Tokyo. 152-8550, Japan (e-mail:
zliu@ok.sc.e.titech.ac.jp).}% <-this % stops a space
}

% \thanks{Manuscript received April 19, 2021; revised August 16, 2021.}}

% The paper headers
\markboth{Journal of \LaTeX\ Class Files,~Vol.~14, No.~8, August~2021}%
{Shell \MakeLowercase{\textit{et al.}}: A Sample Article Using IEEEtran.cls for IEEE Journals}

% \IEEEpubid{0000--0000/00\$00.00~\copyright~2021 IEEE}
% Remember, if you use this you must call \IEEEpubidadjcol in the second
% column for its text to clear the IEEEpubid mark.
\maketitle

% Row 1 : Imagevolume　MULTIVIEW
% Row 2　：INSTANCE　MASK
% Row ３：BBOX
% Row ３：DYMAMIC

\begin{abstract}
Monocular 3D object detection is a fundamental yet challenging task in 3D scene understanding. Existing approaches heavily depend on supervised learning with extensive 3D annotations, which are often acquired from LiDAR point clouds through labor-intensive labeling processes. To tackle this problem, we propose VSRD++, a novel weakly supervised framework for monocular 3D object detection that eliminates the reliance on 3D annotations and leverages neural-field-based volumetric rendering with weak 2D supervision. \textbf{VSRD++} consists of a two-stage pipeline: multi-view 3D autolabeling and subsequent monocular 3D detector training. In the multi-view autolabeling stage, object surfaces are represented as signed distance fields (SDFs) and rendered as instance masks via the proposed instance-aware volumetric silhouette rendering. To optimize 3D bounding boxes, we decompose each instance’s SDF into a cuboid SDF and a residual distance field (RDF) that captures deviations from the cuboid. To address the geometry inconsistency commonly observed in volume rendering methods applied to dynamic objects, we model the dynamic
objects by including velocity into bounding box attributes
as well as assigning confidence to each pseudo label. Moreover,
we also employ a 3D attribute initialization module to
initialize the dynamic bounding box parameters. In the monocular 3D object detection phase, the optimized 3D bounding boxes serve as pseudo labels for training monocular 3D object detectors. Extensive experiments on the KITTI-360 dataset demonstrate that VSRD++ significantly outperforms existing weakly supervised approaches for monocular 3D object detection on both static and dynamic scenes. Code is available at \url{https://github.com/Magicboomliu/VSRD_plus_plus}.
\end{abstract}

% Key Words

% Introduction Sections
\section{Introduction}
\label{sec:introduction}

\IEEEPARstart{3}{D} object detection is crucial for autonomous vehicle perception systems, enabling accurate perception and safe interaction with the environment. 
This task involves precisely locating, scaling, and orienting objects in three-dimensional space, primarily using data from LiDAR point clouds, multi-view imagery, and monocular images. Among these, monocular 3D object detection is notably challenging due to the depth ambiguities inherent in monocular settings. Most current approaches in 3D object detection, including methods like FCOS3D~\cite{FCOS3D}, DD3D~\cite{DD3D}, MonoDIS~\cite{MonoDIS}, MonoFlex~\cite{MonoFlex}, and MonoDETR~\cite{MonoDETR}, as well as the emerging techniques by Jinrang et al. \cite{jinrang2024monouni}, are grounded in supervised learning paradigms which heavily depend on large datasets with 3D bounding box and semantic labels annotated based on LiDAR point cloud. This reliance on extensive manual annotations results in significant labor and cost, hindering both the loop processing and scalability of these technologies. Consequently, the high costs associated with manual data annotation restrict the widespread deployment and practical application of these detection systems across varying autonomous driving scenarios.

\begin{figure}[t]
    \centering
    \includegraphics[width=\linewidth]{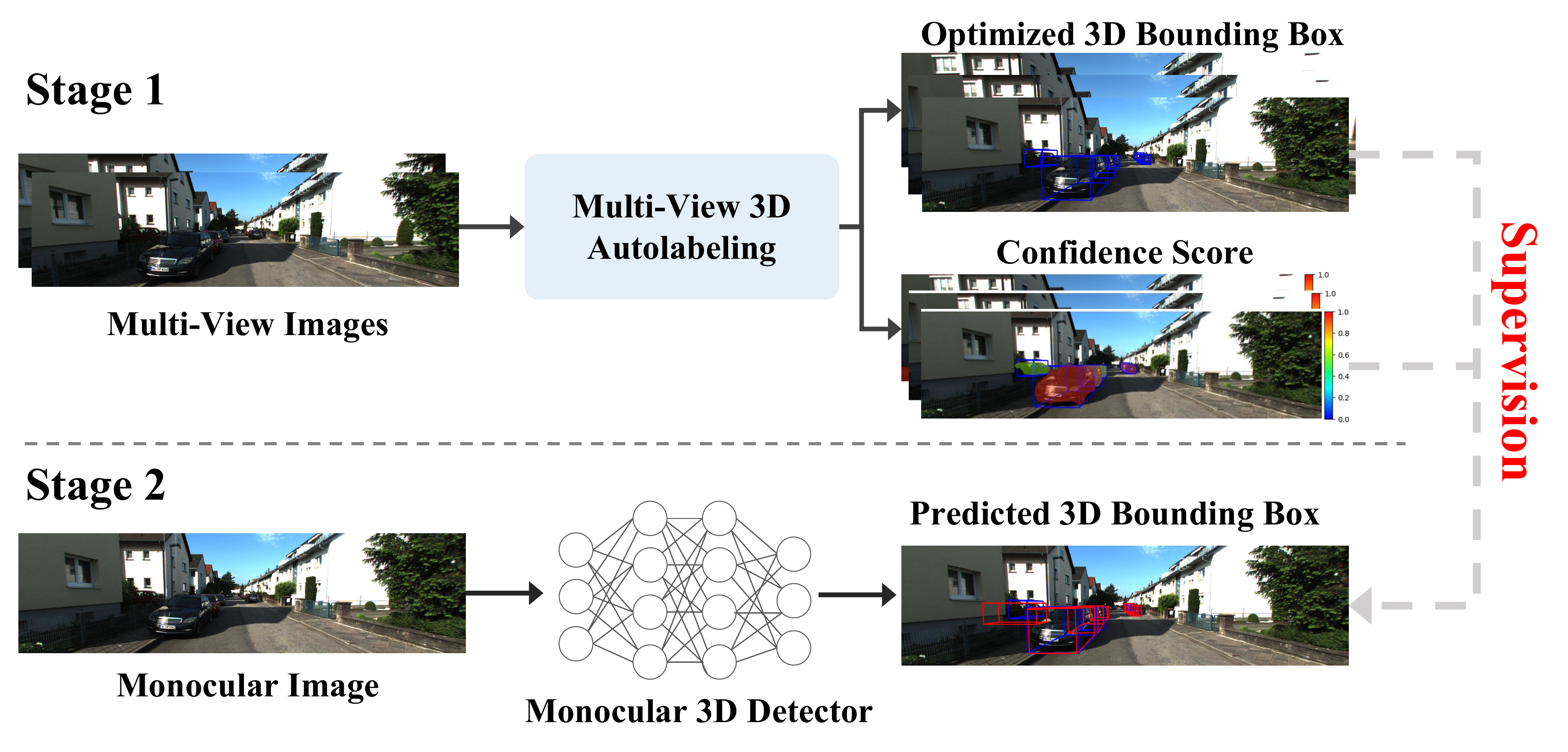}
    \caption{Illustration of our proposed two-stage weakly supervised 3D object detection framework, which consists of multi-view 3D auto-labeling and subsequent training of monocular 3D object detectors using the pseudo labels generated in the auto-labeling stage.}
    \label{fig:overall_framework}
\end{figure}

Although 3D-labeled datasets are scarce and costly to annotate, 2D-labeled datasets like segmentation and instance masks are more abundant and easy to obtain. Furthermore, the acquisition of 2D labels is significantly cheaper than 3D labels. Advanced models such as Segment Anything~\cite{SegmentAnything} also enhance the affordability and accessibility of 2D labels. Recently, neural-based volume rendering techniques such as NeRF~\cite{NeRF} and NeuS~\cite{NeuS} have demonstrated the capability to model 3D scenes from multi-view 2D observations without requiring 3D labels. Inspired by these methods, we developed \textbf{VSRD++} (\underline{\textbf{V}}olumetric \underline{\textbf{S}}ilhouette \underline{\textbf{R}}endering for \underline{\textbf{D}}etection), a novel weakly supervised framework for 3D object detection. This approach facilitates the training of 3D object detectors without reliance on 3D supervision such as 3D bounding boxes and LiDAR point clouds. As illustrated in Figure~\ref{fig:overall_framework}, our proposed VSRD++ framework consists of two stages: an multi-view 3D autolabeling stage that generates pseudo labels, followed by a training stage where these labels are used to train a monocular 3D object detector.   

In the multi-view 3D autolabeling stage, we approach the 3D detection task as a surface 3D reconstruction task. Specifically, we represent each instance's surface by employing a 3D bounding box to approximate its shape, along with a residual that reflects the spatial gap between the bounding box surface and the actual object surfaces. Inspired by NeuS~\cite{NeuS}, we employ the signed distance field (SDF) to implicitly represent the surface and render its silhouette as an instance mask through volumetric rendering. By comparing the rendered instance masks with the ground-truth instance masks, we can refine the SDF and subsequently enhance the optimization of the 3D bounding boxes. 

There are two key mechanisms in this multi-view autolabeling stage. The first is the SDF decomposition, whereby the SDF of each instance is decomposed into the SDF of a cuboid and the \textit{residual distance field} (RDF) that represents the residual from the cuboid. This decomposition models the spatial gap between the surfaces of each instance and the 3D bounding box, enabling more accurate silhouette rendering and providing more reliable feedback signals during optimization. The second key mechanism is the instance-aware volumetric silhouette rendering, which integrates instance labels along a ray to specifically render the silhouette of each individual instance rather than the entire scene. This approach allows for the rendering of each instance’s silhouette while accounting for geometric relationships among instances, such as occlusion. 

Besides, in multi-view volumetric rendering, accurate 3D shape reconstruction relies on maintaining geometric consistency across viewpoints. However, dynamic objects change positions over time, complicating the accurate derivation of SDF. This variability can lead to poor performance in optimized bounding boxes in highly dynamic environments. To enhance the modeling of dynamic objects, we expand the attributes of the source view instance's 3D bounding box by incorporating velocity as an additional optimizable attribute. This modification extends the bounding box from a 7-DoF to a 10-DoF dynamic bounding box. Moreover, We also employ a 3D initialization module to initialize the 10-DoF bounding box parameters, experimental results show that this attribute initialization not only stabilizes the optimization process but also aids in modeling dynamic objects by providing an initial state for optimization and a dynamic mask to specify dynamic objects.

In stage 2, the 3D bounding boxes optimized through our autolabeling process serve as pseudo labels for training 3D object detectors. However, challenges such as dynamic objects with complicated motion that cannot be modeled by a constant velocity and inaccurate camera poses can degrade the quality of these pseudo labels, adversely impacting the training outcomes. To mitigate this issue, we propose a straightforward yet effective algorithm that assigns a confidence score to each pseudo label, reflecting its reliability. Our experiments demonstrate that integrating these confidence scores as per-instance loss weights enhances the performance of 3D object detectors trained with these pseudo labels.

In summary, our main contributions are as follows:
\begin{itemize}
  \item We propose a weakly supervised 3D object detection framework called VSRD++ consisting of multi-view 3D autolabeling and subsequent training of monocular 3D object detectors.
  \item We introduce a novel instance-aware volumetric silhouette rendering method that enables the rendering of each instance’s silhouette as an instance mask.
  \item We present a novel SDF decomposition technique where each instance’s SDF is decomposed into the SDF of a cuboid and a residual distance field (RDF).
  \item To handle dynamic objects, we introduce two effective methods for the multi-view 3D autolabeling and monocular 3D detector training stages. We model the dynamic objects by including velocity into bounding box attributes as well as assigning confidence to each pseudo label.
  \item We propose a 3D attribute initialization pipeline that leverages self-supervised pseudo-depth to obtain initial attributes for 3D bounding boxes, increasing the stability of the optimization process.
  \item Extensive experiments on the KITTI-360 dataset demonstrate that our method outperforms existing weakly supervised 3D object detection methods.
\end{itemize}

% Fixed Here
This paper is an extended version of our conference paper~\cite{liu2024vsrd}. Compared with VSRD~\cite{liu2024vsrd}, our VSRD++ extends the 3D bounding boxes with an explicit velocity attribute and builds a time-varying SDF to model the dynamic objects, detailed in Section~\ref{sec:method:handling_of_dynamic_objects:time-dependent-box-residual-via-velocity-modeling}. Moreover, we further propose a novel 3D attribute initialization method in Section~\ref{sec:method:handling_of_dynamic_objects:self-supervised-3d-attribute-initialzation}, which leverages unsupervised depth information to efficiently initialize 3D bounding boxes and velocities. Additionally, we extend the ablation studies in VSRD++ to comprehensively evaluate all existing and newly introduced modules in Section~\ref{sec:experiments/ablation_study}, we also conducted new comparative experiments in Section~\ref{sec:experiments/evaluation_results} in both static scenes and dynamic scenes, demonstrating that VSRD++ significantly outperforms the original VSRD framework in both accuracy and robustness. These results underscore VSRD++ as a promising approach for advancing weakly supervised 3D object detection for various scenarios.

% Related Works
\section{Related Work}
\label{sec:related_work}

\subsection{Monocular 3D Object Detection}
\label{sec:related_work/monocular_3d_object_detction}
Monocular 3D detection is a challenging task due to limited 3D information from monocular imagery. Deep3DBox \cite{DeepBox3D} pioneered this area by regressing relatively stable 3D object properties and combining these estimates with geometric constraints provided by the 2D bounding box. FCOS3D \cite{FCOS3D} employs a fully convolutional single-stage detector, transforming 7-DoF 3D targets to the image domain and decoupling them as 2D and 3D attributes. To enhance the performance of monocular 3D object detection for truncated objects, MonoFlex \cite{MonoFlex} explicitly decouples truncated objects and adaptively combines multiple approaches for object depth estimation. M3D-RPN \cite{M3D-PRN} and MonoDETR \cite{MonoDETR} also explored the usage of depth cues to improve monocular 3D object detection. The former designed depth-aware convolutional layers that enable location-specific feature extraction and consequently improved 3D scene understanding, while the latter modified the vanilla Transformer \cite{ViT} to be depth-aware to guide the whole detection process by contextual depth cues. Despite these advances, the reliance on expensive and labor-intensive manual annotation on LiDAR point clouds remains a significant limitation.  

\subsection{Weakly Supervised 3D Object Detection}
\label{sec:related_work/weakly_supervised_3d_object_detection}
Many weakly supervised methods have been proposed to mitigate the high annotation cost in 3D object detection. WS3D \cite{W3D} introduced a LiDAR-based two-stage pipeline where cylindrical object proposals are first generated under weak supervision and then refined using a few labeled object instances. SMOKE \cite{SMOKE} estimates 3D bounding boxes by combining keypoint estimates with regressed 3D box parameters. FCOS3D \cite{FCOS3D} employs a fully convolutional single-stage detector, transforming 7-DoF 3D targets to the image domain and decoupling them as 2D and 3D attributes. VS3D \cite{VS3D} introduced an unsupervised 3D proposal module that generates object proposals by leveraging normalized point cloud densities. WeakM3D \cite{WeakM3D} introduced a weakly supervised monocular 3D object detection method that leverages the 3D alignment loss between each predicted 3D bounding box and corresponding RoI LiDAR points. Furthermore, it introduced a method to estimate the orientation from RoI LiDAR points based on its statistics. Recently, WeakMono3D \cite{WeakMono3D} leverages projection loss with multi-view and direction consistency, achieving a weakly supervised monocular object detection that relies only on 2D supervision. However, its reliance on 2D direction annotations restricts its applicability to large-scale datasets. Zakharov et al. \cite{Autolabels} proposed an autolabeling pipeline that integrates an SDF-based differentiable shape renderer and normalized object coordinate spaces (NOCS). However, additional training on synthetic data is still required for shape and coordinate estimation. In contrast, our method purely relies on 2D supervision, eliminating the necessity of synthetic data or 3D supervision.

\subsection{3D Object Detection with Neural Fields}
\label{sec:related_work/3d_object_detection_with_neural_fields}
Neural Radiance Fields (NeRF) \cite{NeRF} introduced a new perspective for implicit learning of 3D geometry from posed multi-view images by volume rendering. Building upon the vanilla NeRF, subsequent research has focused on enhancing novel view synthesis \cite{RegNeRF, MipNeRF, MipNeRF360, ZipNeRF} or accelerating volume rendering \cite{InstantNGP, Plenoxels, TensoRF}. NeuS \cite{NeuS} introduced a novel volume rendering scheme to learn a neural SDF representation by introducing a density distribution induced by the SDF. Similarly, VolSDF \cite{VolSDF} defined the volume density function as Laplace’s cumulative distribution function applied to an SDF representation. Many works \cite{NeRFDet, NeRF-RPN, MonoNeRD} recently have attempted to utilize neural fields for 3D object detection. Notably, NeRF-RPN \cite{NeRF-RPN} demonstrated that the 3D bounding boxes of objects in NeRF can be directly regressed without rendering. NeRF-Det \cite{NeRFDet} connects the detection and NeRF branches through a shared MLP, enabling an efficient adaptation of NeRF to detection and yielding geometry-aware volumetric representations. MonoNeRD \cite{MonoNeRD} models scenes with SDFs and renders images and depth through volume rendering to obtain intermediate 3D representations for detection. However, these approaches still rely on ground truth 3D labels for supervision. In contrast, we propose the first volume rendering-based weakly supervised 3D object detection framework that relies on multi-view geometry and 2D supervision without any 3D supervision, such as 3D bounding boxes or LiDAR point clouds.

\subsection{Neural Rendering for Dynamic Scenes}\label{sec:related_work/neural_rendering_for_dyanmic}
Building radiance fields for dynamic environments marks a pivotal progression in Neural Radiance Field (NeRF), offering considerable potential for practical deployments. D-NeRF~\cite{D-NeRF}, Nerfies~\cite{nerfies} and HumanNeRF~\cite{HumanNeRF} modify the standard NeRF framework to handle dynamic settings by incorporating deformation fields. In contrast, DyNeRF~\cite{DyNeRF} models intricate temporal changes by mapping a specific latent vector to each frame. 
HexPlane~\cite{cao2023hexplane} utilizes planar factorization to extend into arbitrary dimensions, enabling dynamic scene representation. As for SDF-Based Volume Rendering, Neus2~\cite{wang2023neus2} advances NeuS~\cite{NeuS} by constructing the initial frame using static reconstruction and subsequently predicting and accumulating global transformations for each frame to align with the canonical space. Zhang et al.~\cite{MeshMM24} achieve dynamic mesh reconstruction by optimizing Mesh-Centric SDF with Gaussian Splatting, effectively bypassing the slower Marching Cubes process and yielding impressive results. While these approaches demonstrate strong performance in modeling dynamic objects, they have two notable limitations: (1) they require time-intensive multi-view observations to construct deformable fields, and (2) they rely on additional MLPs or parameters for dense offset modeling. Given the sparse nature of 3D detection and predominantly rigid object motions, more efficient approaches are needed.

\subsection{Open-vocabulary 3D object detection.}
\label{sec:related_works:open_voc}
Open-vocabulary 3D object detection (OV3DDet) aims to detect 3D objects beyond a fixed set of training categories by leveraging vision--language pre-training and cross-modal alignment.
Recent works such as CoDA~\cite{Cao2023CoDA} collaboratively discover novel 3D boxes and align them with text embeddings for open-vocabulary 3D detection, while Jiao et al.~\cite{Jiao2024OV3DDet} further enhance open-vocabulary 3D detection by jointly exploiting textual and visual guidance.
More recently, Cao et al.~\cite{Cao2025TPAMI_OV3DDet} extend this line of research with a TPAMI version that improves collaborative novel object discovery and box-guided cross-modal alignment.
In contrast to these methods, VSRD++ focuses on closed-set, weakly supervised monocular 3D object detection on driving scenes: we assume a fixed category set (e.g., \textit{Car}) and aim to eliminate the need for 3D annotations by volumetric autolabeling from multi-view images.
Our instance-aware volumetric silhouette rendering and SDF-based pseudo labeling are complementary to OV3DDet: in principle, open-vocabulary 2D or 3D proposals from~\cite{Cao2023CoDA,Jiao2024OV3DDet,Cao2025TPAMI_OV3DDet} could be combined with our volumetric optimization to generate category-agnostic 3D pseudo labels for novel classes, which we leave as an interesting direction for future work.

% Method
\section{Method}
\label{sec:method}

\begin{figure*}[t]
   \scalebox{0.95}{
    \centering
 
\includegraphics[width=\linewidth]{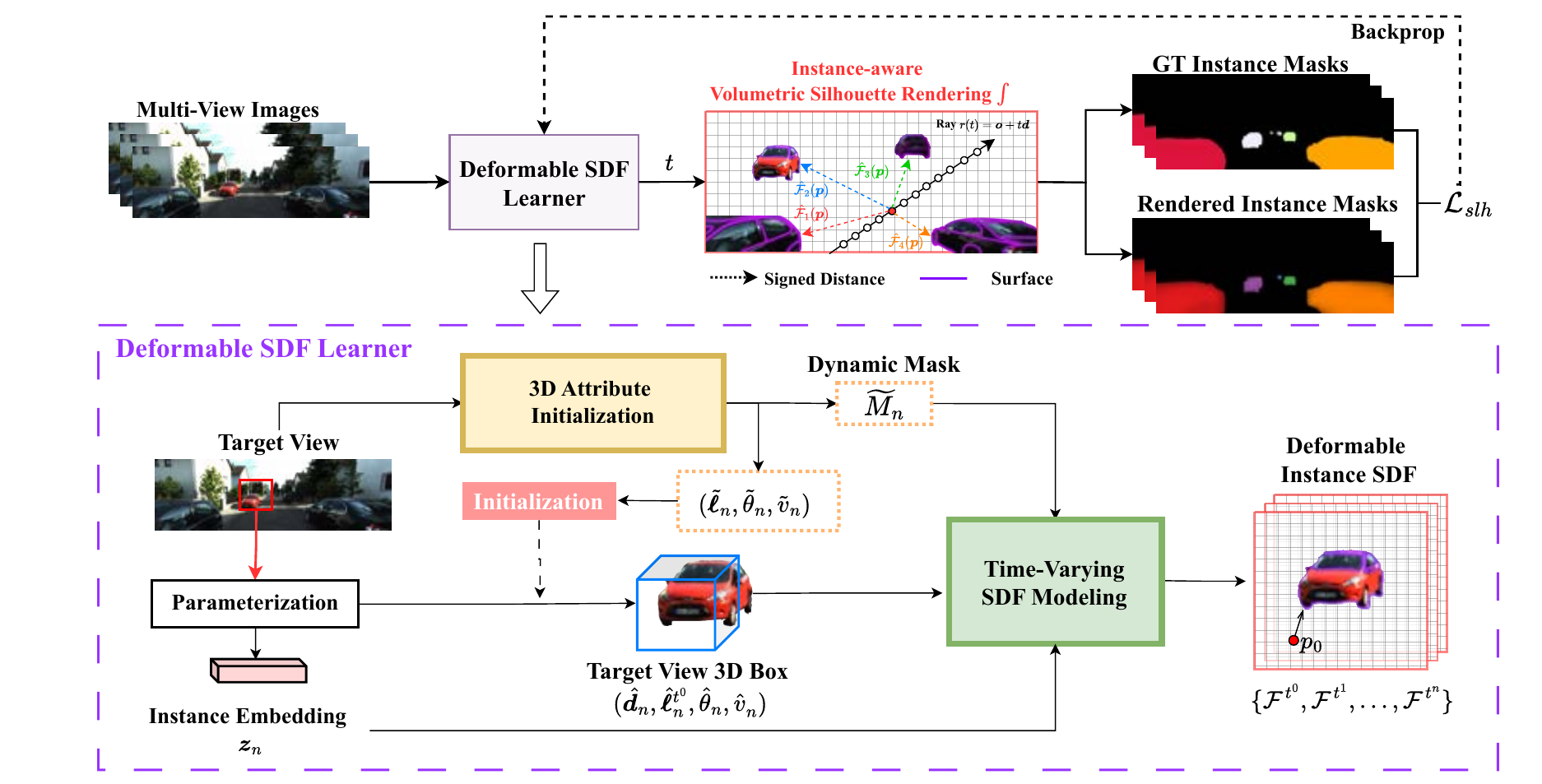}}
    \caption{Illustration of the pipeline for our proposed multi-view 3D auto-labeling framework, \textbf{VSRD++}. Each instance surface is represented as a signed distance field (SDF) and optimized through a \textit{Deformable SDF Learner}. To initialize deformable 3D bounding box attributes (e.g., location, orientation, velocity), we employ the  3D attribute initialization module. The time-variant SDF modeling incorporates velocity to decouple instance SDF dynamics. The composed instance SDF enables silhouette rendering via \textit{Instance-Aware Volumetric Silhouette Rendering}. All 3D bounding boxes are optimized by minimizing the loss between the rendered and ground truth instance masks.}
    \label{fig:method/vsrd_multi_view_3d_auto_labeling}
    
\end{figure*}

\subsection{Overview}
\label{proposed_overall}
As illustrated in Figure~\ref{fig:method/vsrd_multi_view_3d_auto_labeling}, our proposed \textbf{VSRD++} is a two-stage, weakly supervised 3D object detection framework comprising a multi-view 3D autolabeling followed by the training of monocular 3D object detectors using the generated pseudo labels. In this section, we first provide a brief introduction to volume rendering, specifically SDF-based volume rendering for surface reconstruction, a pivotal technique utilized in our VSRD++ pipeline, in Section~\ref{sec:method:perliminaries}. A comprehensive overview of the multi-view 3D autolabeling pipeline is presented in Section~\ref{sec:method/multi_view_3d_auto_labeling_old}. Furthermore, to address the challenges of modeling dynamic objects and optimizing 3D bounding boxes, we introduce two novel modules in Section~\ref{sec:method:handling_of_dynamic_objects} of our framework. Detailed strategies for training the 3D detectors with pseudo labels are discussed in Section~\ref{sec:method/training_of_3d_object_detectors}. Lastly, the loss functions utilized in both the multi-view 3D autolabeling phase and the training of the monocular 3D object detectors are described in Section~\ref{sec:method:loss_functions}.

% Preliminaries SDF Modeling
\subsection{Preliminaries}\label{sec:method:perliminaries}
\paragraph{SDF-based Volumetric Rendering}
\label{sec:method/multi_view_3d_auto_labeling/preliminaries/sdf_based_volumetric_rendering}
NeRF~\cite{NeRF} represents a 3D scene with neural density and color fields. Given a camera position $\bm{o} \in \mathbb{R}^{3}$ and a ray direction $\bm{d} \in \mathbb{R}^{3}$ emitted from a pixel, the volume rendering scheme integrates the colors of sampled points along the ray as follows:
\begin{align}
    \label{eq:method/multi_view_3d_auto_labeling/preliminaries/sdf_based_volumetric_rendering/integration}
    \hat{\bm{C}}(\bm{o}, \bm{d}) & = \int_{0}^{\infty} w(t) \bm{c}(\bm{r}(t), \bm{d}) \mathrm{d}t \ , \\
    \label{eq:method/multi_view_3d_auto_labeling/preliminaries/sdf_based_volumetric_rendering/volume_density_weight}
    w(t) & = \exp\left(- \int_{0}^{t} \sigma(\bm{r}(u)) \mathrm{d}u\right) \sigma(\bm{r}(t)) \ , 
\end{align}
where $\hat{\bm{C}}(\bm{o}, \bm{d}) \in \mathbb{R}^{3}$ denotes the rendered color of the pixel, $\bm{r}(t) = \bm{o} + t\bm{d}$ denotes the ray, $\bm{c}(\bm{p}, \bm{d})$ denotes the color at position $\bm{p}$ and view direction $\bm{d}$, and $\sigma(\bm{p})$ denotes the \textit{volume density} at position $\bm{p}$. 

Since the density field cannot explicitly represent surfaces, NeuS~\cite{NeuS} introduced an SDF-based volume rendering formulation, where a surface represented by an SDF is reinterpreted as a participating medium modeled by a density field, allowing surface rendering through volume rendering. In this formulation, the weight $w(t)$ in Eq.~(\ref{eq:method/multi_view_3d_auto_labeling/preliminaries/sdf_based_volumetric_rendering/volume_density_weight}) is redefined by introducing \textit{opaque density} $\rho(t)$ as follows:
\begin{align}\label{eq:method/multi_view_3d_auto_labeling/preliminaries/sdf_based_volumetric_rendering/opaque_density_weight}
    w(t) & = \exp\left(- \int_{0}^{t} \rho(u) \mathrm{d}u\right) \rho(t) \ , \\
\label{eq:method/multi_view_3d_auto_labeling/preliminaries/sdf_based_volumetric_rendering/opaque_density}
    \rho(t) & = \max\left(- \frac{\frac{\mathrm{d} \Phi}{\mathrm{d} t} (\hat{\mathcal{F}}(\bm{r}(t)))}{\Phi(\hat{\mathcal{F}}(\bm{r}(t)))}, 0\right) \ ,
\end{align}
where $\hat{\mathcal{F}}(\cdot)$ denotes the SDF for the entire scene and $\Phi(\cdot)$ represents the Sigmoid function. Our proposed instance-aware volumetric silhouette rendering, introduced in Section~\ref{sec:method/multi_view_3d_auto_labeling_old/instance_aware_volumetric_silhoutette}, follows the same SDF-based weight formulation as Eq.~(\ref{eq:method/multi_view_3d_auto_labeling/preliminaries/sdf_based_volumetric_rendering/opaque_density_weight}) but incorporates instance labels instead of colors along a ray to render instance masks.

% Multi-View 3D Auto-Labeling Phase
\subsection{Multi-View 3D Autolabeling Based on Instance-Aware Volumetric Silhouette Rendering}\label{sec:method/multi_view_3d_auto_labeling_old} 
In this subsection, we introduce our multi-view 3D autolabeling pipeline, which transforms the task of bounding box optimization into an SDF-based surface reconstruction task, enabling the use of volume rendering to derive 3D bounding boxes solely from 2D observational data. As shown in Figure~\ref{fig:method/vsrd_multi_view_3d_auto_labeling}, our proposed method processes sequential images as input and constructs a \textit{deformable SDF learner} for 3D bounding box optimization. This learner captures the instance-specific SDFs of the scene, followed by instance-aware volumetric silhouette rendering to generate the rendered instance masks for loss backpropagation. We outline the problem definition in Section~\ref{sec:method:multi_view:problem_definition}, elaborate on integrating bounding boxes with instance-specific SDFs in Sections~\ref{sec:method/multi_view_3d_auto_labeling_old/3d_box_as_sdf} and~\ref{sec:method/multi_view_3d_auto_labeling_old/residual_distance_field}, and introduce instance-aware volumetric rendering for instance masks computation.

% SDF Decomposition Illustration
% SDF Decomposition
\begin{figure}[!t]
    \centering
    \includegraphics[width=\linewidth]{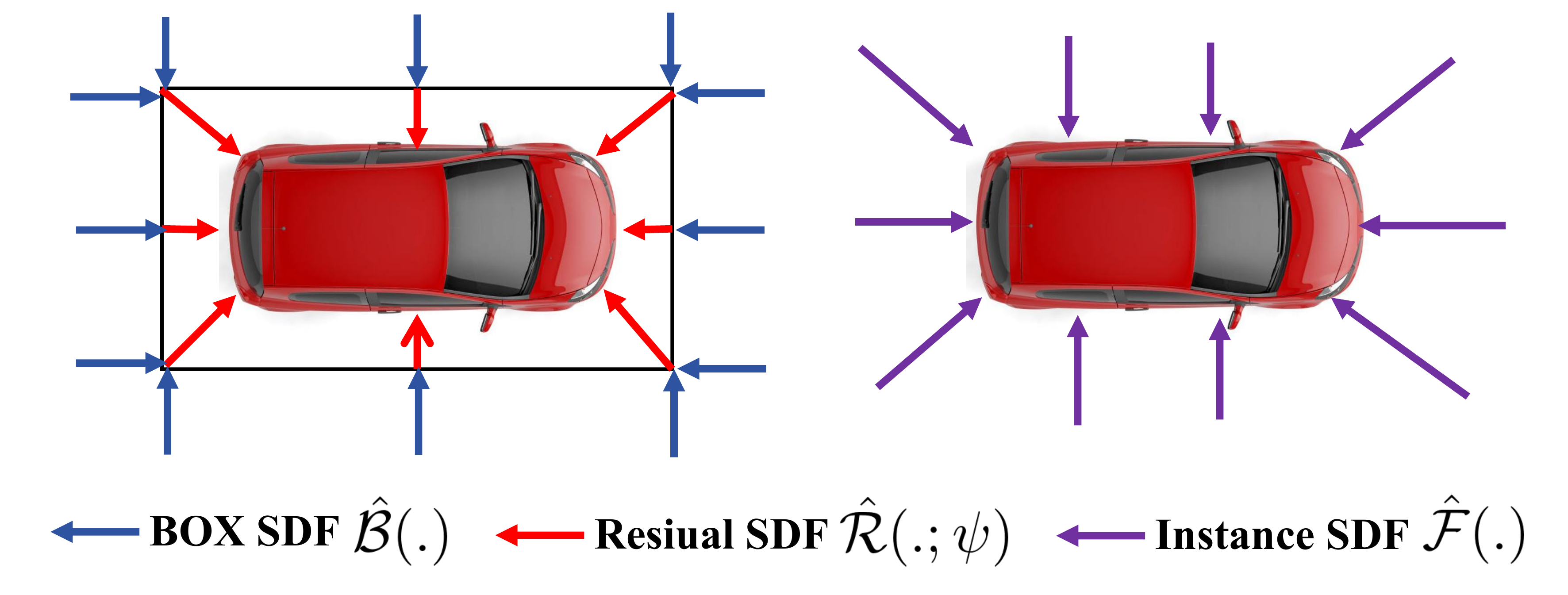}
    \caption{Illustration of the instance SDF decomposition, where we decouple the surface of the cars into the combination of the cuboid box SDF which is represented by the 3D bounding boxes and the spatial residual from the cuboid. \textcolor[rgb]{0,0,1}{Blue arrow} presented for the cuboid box SDF parameterized by the 3D bounding boxes, where the \textcolor[rgb]{1,0,0}{red arrow} demonstrate the residual RDF and the \textcolor[rgb]{0.5,0,1}{purple arrow} represents the instance SDF.  }
    \label{fig:BoudningBox_Decouple}
\end{figure}
\subsubsection{Problem Definition}
\label{sec:method:multi_view:problem_definition}
Given a monocular video consisting of posed frames with annotated instance masks, our goal is to optimize the 3D bounding boxes of each frame without utilizing ground truth 3D bounding boxes or LiDAR point clouds for supervision. More specifically, for each \textit{target} frame $t$ in the video sequence, we sample multiple \textit{source} frames $\mathcal{S}$ and optimize the $N$ 3D bounding boxes in the target frame using the instance masks of the source frames as the weak supervision, where $N$ denotes the number of instances in the target frame. 

We parameterize the $n$-th 3D bounding box $\hat{\bm{B}}_{n} \in \mathbb{R}^{7 \times 3}$ in the target frame with a dimension $\hat{\bm{d}}_{n} \in \mathbb{R}^{3}_{+}$, location $\hat{\bm{\ell}}_{n} \in \mathbb{R}^{3}$, and orientation $\hat{\theta}_{n} \in \mathbb{R}$. In addition to these parameters for each bounding box, we introduce a learnable instance embedding $\bm{z}_{n} \in \mathbb{R}^{D}$ for each instance and a shared \textit{hypernetwork} parameterized by $\bm{\psi}$ for the \textit{residual distance field} introduced in Section~\ref{sec:method/multi_view_3d_auto_labeling_old/residual_distance_field}. We stack each parameter group into a single tensor over all the instances, yielding dimensions $\hat{\bm{D}} \in \mathbb{R}^{N \times 3}_{+}$, locations $\hat{\bm{L}} \in \mathbb{R}^{N \times 3}$, orientations $\hat{\bm{\Theta}} \in \mathbb{R}^{N \times 1}$, and instance embeddings $\bm{Z} \in \mathbb{R}^{N \times D}$. 

Given the loss function $\mathcal{L}$ explained in Section~\ref{sec:method:loss_functions_for_multi_view_3D_autolabelsing}, and considering the presence of dynamic objects, we extend the 3D bounding box into a 10-DoF bounding box $\hat{\bm{B}}_{n} \in \mathbb{R}^{10 \times 3}$ with an additional learnable velocity attribute, represented by $\hat{\bm{V}} \in \mathbb{R}^{N \times 3}$, to model time-varying locations. For more details on handling dynamic objects, please refer to Section~\ref{sec:method:handling_of_dynamic_objects}. Finally, the optimized 3D bounding boxes $^{*}\!{\bm{B}}$ decoded from $^{*}\!\bm{D}$, $^{*}\!\bm{L}$, $^{*}\!\bm{\Theta}$, and $^{*}\!\bm{V}$ can be used as pseudo labels for training 3D object detectors, as explained in Section~\ref{sec:method/training_of_3d_object_detectors}.

\begin{figure}[t]
    \centering
    \includegraphics[width=\linewidth]{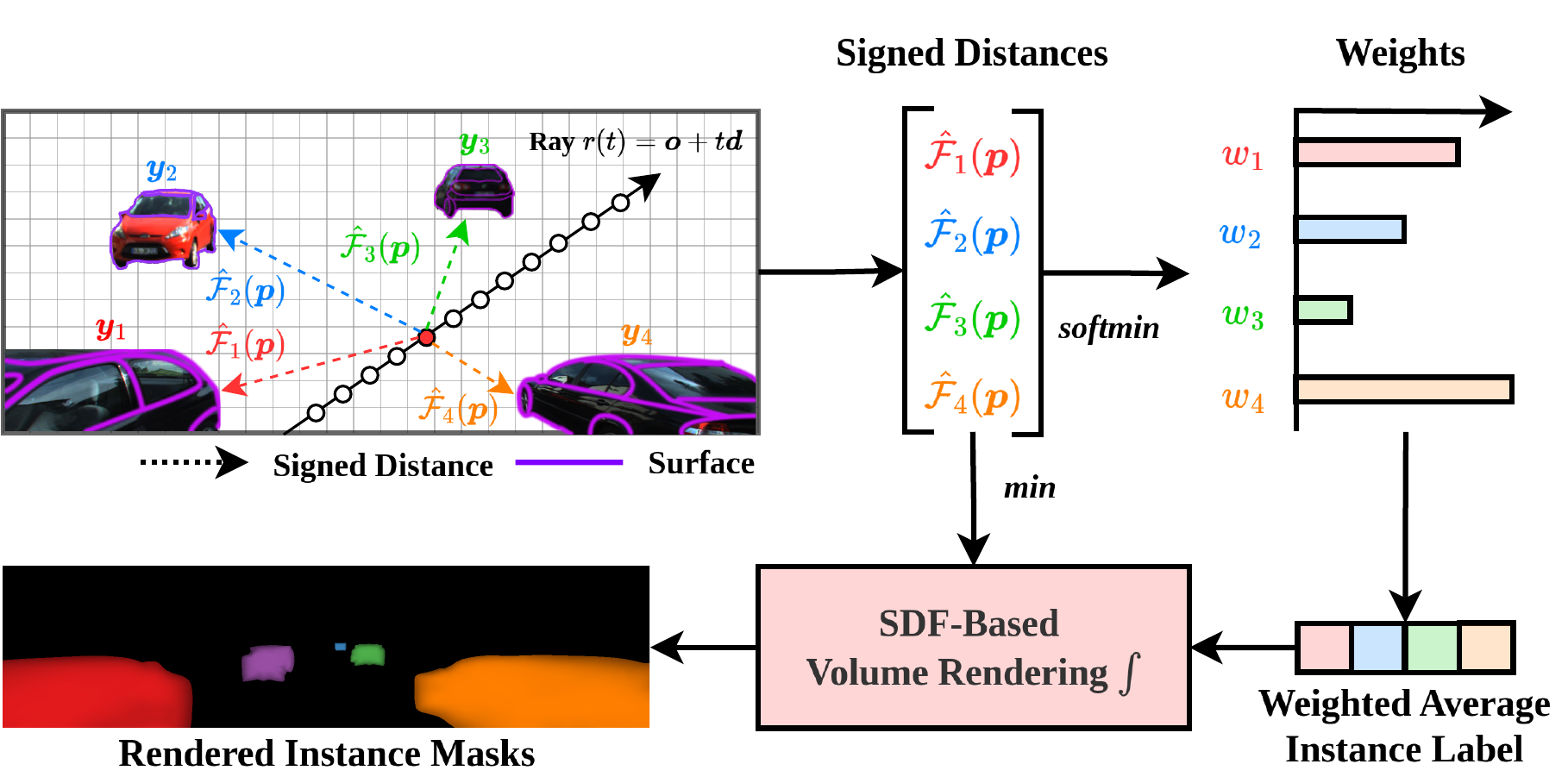}
    \caption{Illustration of our proposed instance-aware volumetric silhouette rendering. The instance labels are averaged for each sampled point along a ray based on the signed distance to each instance. The averaged instance labels are integrated along the ray based on the SDF-based volume rendering formulation \cite{NeuS}.}
    \label{fig:method/multi_view_3d_auto_labeling/instance_aware_volumetric_silhouette_rendering}
\end{figure}

\subsubsection{3D Bounding Box Represented as an SDF}\label{sec:method/multi_view_3d_auto_labeling_old/3d_box_as_sdf} 
To optimize the 3D bounding boxes in each target frame through volume rendering, we first represent the surface of each 3D bounding box as a signed distance field (SDF), which is one of the most common surface representations. An SDF is defined as a function $\mathcal{F} : \mathbb{R}^{3} \rightarrow \mathbb{R}$ that maps a spatial position $\bm{p} \in \mathbb{R}^{3}$ to its signed distance to the closest point on the surface. The zero-level set $\{\bm{p} \in \mathbb{R}^{3} \mid \mathcal{F}(\bm{p}) = 0\}$ represents the surface itself. The SDF for a cuboid, parameterized by its dimension $\bm{d} \in \mathbb{R}^{3}_{+}$, location $\bm{\ell} \in \mathbb{R}^{3}$, and orientation $\bm{R} \in \text{SO}(3)$, can be derived theoretically and is denoted by $\mathcal{B}(\cdot ; \bm{d}, \bm{\ell}, \bm{R})$. This transformation enables the direct conversion of 3D bounding box parameters into SDFs, facilitating volume rendering and optimization.

\subsubsection{Residual Distance Field (RDF)}\label{sec:method/multi_view_3d_auto_labeling_old/residual_distance_field} 
In general, the shape of each instance is not a perfect cuboid. If we rely only on the cuboid SDF introduced in Section~\ref{sec:method/multi_view_3d_auto_labeling_old/3d_box_as_sdf} to render the surface of each instance, the spatial gap between the surfaces of the instance and cuboid would lead to inaccurate silhouettes, generating unreliable feedback signals during optimization. To address this, we propose a novel neural field named \textit{residual distance field} (RDF), which models the residual difference between the signed distances to the surfaces of the instance and the 3D bounding box. As illustrated in Figure~\ref{fig:BoudningBox_Decouple}, let $\mathcal{F}_{n}(\cdot)$ be the \textit{true} SDF of the surface enclosed by the $n$-th 3D bounding box $\hat{\bm{B}}_{n}$, whose SDF is given by $\hat{\mathcal{B}}_{n}(\cdot) = \mathcal{B}(\cdot ; \hat{\bm{d}}_{n}, \hat{\bm{\ell}}_{n}, \bm{R}_{y}(\hat{\theta}_{n}))$, where $\bm{R}_{y}(\theta)$ denotes the rotation matrix around the $y$-axis by an angle $\theta$. For any point $\bm{p} \in \mathbb{R}^{3}$, we define the RDF as:
\begin{equation}
\hat{\mathcal{R}}_{n}(\bm{p}) \coloneqq \mathcal{F}_{n}(\bm{p}) - \hat{\mathcal{B}}_{n}(\bm{p}) .
\end{equation}
By definition, since a 3D bounding box encloses the corresponding instance, we require that $\forall \bm{p} \in \mathbb{R}^{3}, \hat{\mathcal{R}}_{n}(\bm{p}) \geq 0$.

A straightforward approach to modeling the RDF for each instance is to train $N$ individual networks. However, since objects within the same semantic class often exhibit similar geometric structures, we employ a single \textit{hypernetwork}~\cite{Hypernetworks} to regress the weights of the neural RDF directly from an instance embedding. Given the $n$-th instance embedding $\bm{z}_{n} \in \mathbb{R}^{D}$, where $D$ denotes the embedding dimension, the $n$-th neural RDF is formulated as:
\begin{align}
\label{eq:method/multi_view_3d_auto_labeling/residual_distance_field/hypernetwork}
    \hat{\mathcal{R}}_{n}(\bm{p}) &= \sigma(\mathcal{G}(\bm{p} ;  \bm{\phi}_{n})) \, , \\
    \bm{\phi}_{n} &= \mathcal{H}(\bm{z}_{n} ; \bm{\psi}) \, 
\end{align}
where $\mathcal{G}(\cdot; \bm{\phi}_{n})$ denotes the $n$-th neural network parameterized by $\bm{\phi}_{n}$, $\mathcal{H}(\cdot; \bm{\psi})$ represents the shared hypernetwork parameterized by $\bm{\psi}$, and $\sigma(\cdot)$ is the Softplus function ensuring the residual distance remains non-negative, consistent with the definition of a bounding box. 

For instance-aware volumetric silhouette rendering, we employ $\hat{\mathcal{F}}_{n}(\bm{p}) = \hat{\mathcal{B}}_{n}(\bm{p}) + \hat{\mathcal{R}}_{n}(\bm{p})$ as the SDF for more accurate silhouette rendering.

\begin{figure*}[!t]
    \centering
    \includegraphics[width=\linewidth]{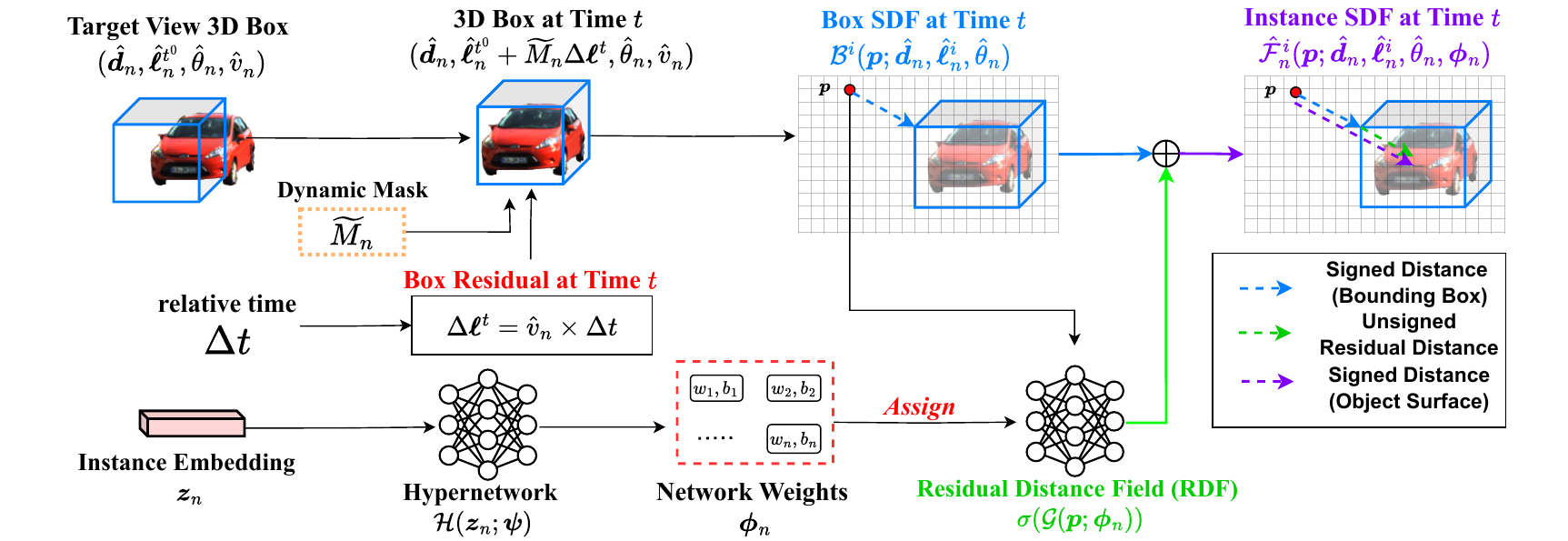}
    \caption{Pipeline of \textbf{Time-Variant SDF Based on  Velocity-Incorporated Bounding Boxes}. We represent the surface of each instance as an SDF and decompose it into the SDF of a 3D bounding box and the residual distance field (RDF), which is learned via a hypernetwork. For each time $t$, we employ relative time duration $\Delta t$ and the velocity $v_{n}$ to obtain the time-dependent box residual to adjust the box SDF field adaptively using the dynamic mask $\tilde{M_{n}}$.}
    \label{fig:velo_deform}
\end{figure*}
\subsubsection{Instance-Aware Volumetric Silhouette Rendering}\label{sec:method/multi_view_3d_auto_labeling_old/instance_aware_volumetric_silhoutette}

After obtaining all the instance SDFs of the given scene, which consists of the cuboid SDF and RDF described in Section~\ref{sec:method/multi_view_3d_auto_labeling_old/3d_box_as_sdf} and Section~\ref{sec:method/multi_view_3d_auto_labeling_old/residual_distance_field}, we further optimize the 3D bounding boxes through volumetric rendering to generate instance masks and compare them with the ground-truth instance masks. To achieve this, we propose a novel SDF-based \textit{instance-aware} volumetric silhouette rendering, where instance labels, rather than colors, are integrated along a ray using the same SDF-based volume rendering formulation as Eq.~(\ref{eq:method/multi_view_3d_auto_labeling/preliminaries/sdf_based_volumetric_rendering/opaque_density_weight}):

\begin{align}
    \label{eq:method/multi_view_3d_auto_labeling/instance_aware_volumetric_silhouette_rendering/integration}
    \hat{\bm{S}}(\bm{o}, \bm{d}) & = \int_{0}^{\infty} w(t) \bm{s}(\bm{r}(t)) dt \, , \\
    \label{eq:method/multi_view_3d_auto_labeling/instance_aware_volumetric_silhouette_rendering/average_instance_label}
    \bm{s}(\bm{p}) & = \sum_{n=1}^{N} \text{softmin}([\hat{\mathcal{F}}_{n}(\bm{p})]_{n=1}^{N})_{n} \cdot \bm{y}_{n} \, ,
\end{align}
where $\hat{\bm{S}}(\bm{o}, \bm{d}) \in [0, 1]^{N}$ denotes the rendered soft instance label, $\bm{s}(\bm{p}) \in [0, 1]^{N}$ represents the weighted average instance label at position $\bm{p}$, indicating the relative proximity of $\bm{p}$ to each instance, and $\bm{y}_{n} \in \{0, 1\}^{N}$ denotes the one-hot instance label of the $n$-th instance. $\hat{\mathcal{F}}_{n}(\cdot)$ represents the SDF of the $n$-th instance, as introduced in Section~\ref{sec:method/multi_view_3d_auto_labeling_old/residual_distance_field}. The weight function $w(t)$ in Eq.~(\ref{eq:method/multi_view_3d_auto_labeling/instance_aware_volumetric_silhouette_rendering/integration}) is derived from Eq.~(\ref{eq:method/multi_view_3d_auto_labeling/preliminaries/sdf_based_volumetric_rendering/opaque_density_weight}). 

To compute Eq.~(\ref{eq:method/multi_view_3d_auto_labeling/preliminaries/sdf_based_volumetric_rendering/opaque_density_weight}), we model the entire scene as the union of all object surfaces, i.e.,
\begin{equation}
    \hat{\mathcal{F}}(\bm{p}) = \min(\hat{\mathcal{F}}_{1}(\bm{p}), \ldots, \hat{\mathcal{F}}_{N}(\bm{p})).
\end{equation}
This mechanism enables the rendering of instance masks while considering geometric relationships among instances, such as occlusions. The entire process is illustrated in Figure~\ref{fig:method/multi_view_3d_auto_labeling/instance_aware_volumetric_silhouette_rendering}.

\subsection{Handling of Dynamic Objects}
\label{sec:method:handling_of_dynamic_objects}
Driving scenes often contain numerous moving objects that disrupt consistency in sequential views at varying locations, degrading the quality of optimized bounding boxes generated by multi-view 3D autolabeling, as shown in Figure~\ref{fig:method/training_of_3d_object_detectors/confidence_assignment}. To enhance the modeling of dynamic objects, we introduce two new modules into our multi-view 3D autolabeling framework: time-varying SDF based on velocity-incorporated bounding boxes and 3D attribute initialization.

\subsubsection{Time-Varying SDF}\label{sec:method:handling_of_dynamic_objects:time-dependent-box-residual-via-velocity-modeling}
To improve our multi-view 3D autolabeling capabilities, we extend 3D bounding boxes to time-varying 3D bounding boxes by incorporating a learnable velocity attribute, allowing more effective modeling of dynamic objects. Figure~\ref{fig:velo_deform} illustrates the overall workflow of our time-varying SDF based on velocity-incorporated bounding boxes. To model dynamic objects, we leverage the relative time duration $\Delta t$ between the current and target frames, along with velocity $v_n$, to compute a time-varying location residual $\Delta L_{n}^{t}$. This allows adaptive adjustment of the 3D bounding box location using the dynamic mask $\tilde{M_{n}}$, formulated as:
\begin{equation}
\hat{L}_{n}^{t} = \hat{L}_{n}^{t_{0}} + \tilde{M_{n}}(\hat{v_{n}} \cdot \Delta t).
\end{equation}
Using this location residual, we extend the original time-shared cuboid SDF into a time-varying SDF that is more flexible and generalizable for dynamic scenes. Moreover, we use the dynamic mask to adaptively control the 3D bounding box modeling. For dynamic objects, time-varying 3D bounding boxes are used for time-varying SDF modeling, whereas for static objects, the SDF remains constant and shared across frames. This mechanism enables more accurate modeling of instance SDFs. Finally, we obtain the instance SDF $\hat{\mathcal{F}_{t}}$ at any given time $t$, which is further used for instance-aware volumetric silhouette rendering. Though this kind of constant velocity is simple, our experiments show substantial improvements in dynamic
regions, as shown in Table~\ref{sec:experiments/ablation_study}, indicating that the current approximation is effective in practice.
\begin{figure*}[!t]
    \centering
    \includegraphics[width=0.93\linewidth]{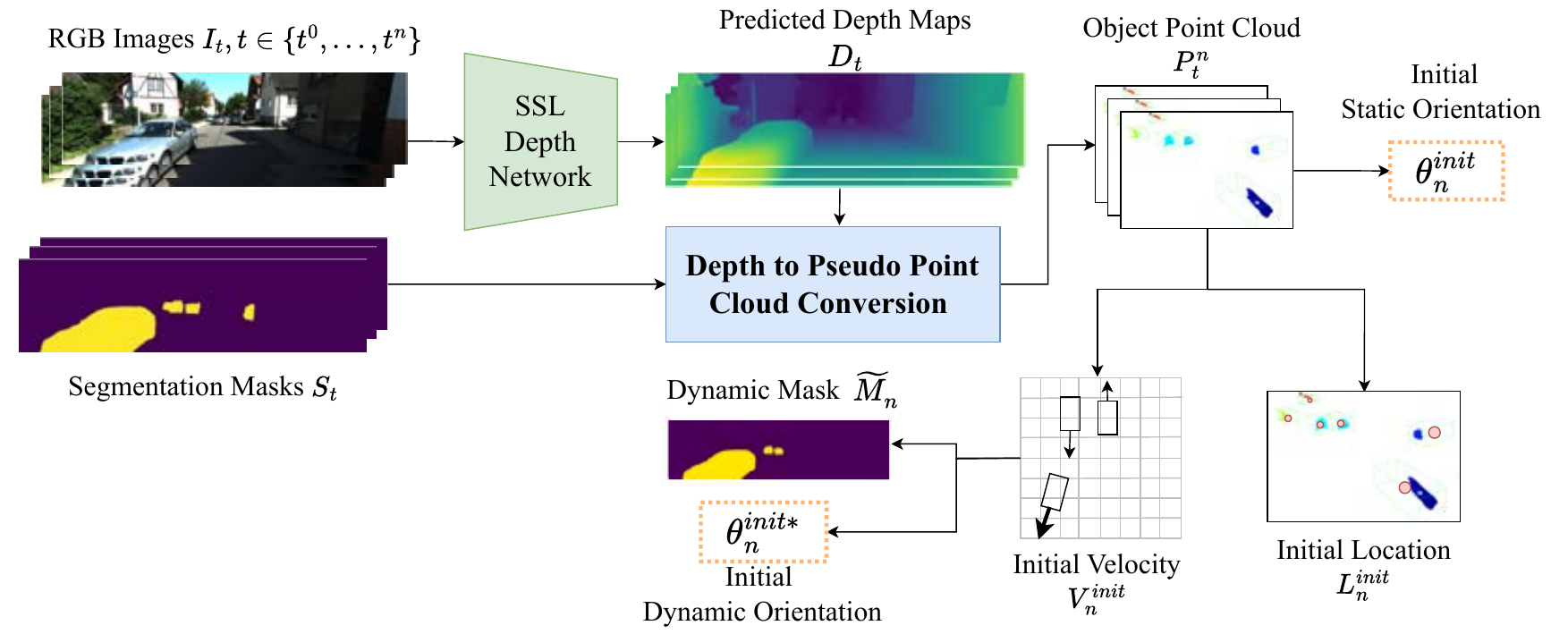}
    \caption{Pipeline of \textbf{3D Attribute Initialization}. Sequential depth maps are generated from RGB inputs using a self-supervised depth estimation network \cite{IGEVStereo} to facilitate the extraction of the initial orientation, location, and velocity of objects, along with dynamic masks for indicating the  object dynamics.}
    \label{fig:Attribute_Initialization}
\end{figure*}
\subsubsection{3D Attribute Initialization}\label{sec:method:handling_of_dynamic_objects:self-supervised-3d-attribute-initialzation}
Volume rendering-based optimization may be unstable and highly sensitive to the initial states of bounding boxes. Additionally, optimization can be hindered by the dynamic nature of objects and the high degrees of freedom when jointly optimizing location and velocity from a random initialization. Therefore, a well-initialized bounding box significantly improves both the stability of the optimization process and the quality of the final results. With the advance of supervised and self-supervised depth estimation for autonomous scenes~\cite{ACVNet,IGEVStereo,NiNet,GOAT,CFDNet,DMS}, we are using pseudo depth for 3D attribute initialization.  As illustrated in Figure~\ref{fig:Attribute_Initialization}, we propose 3D attribute initialization that leverages estimated-depth from a self-supervised depth estimator~\cite{IGEVStereo} to initialize 3D bounding box attributes, including location, orientation, velocity, and a dynamic mask indicating moving objects. Given stereo images $I_{t}^{l}$ and $I_{t}^{r}$ as inputs, we first apply IGEVStereo~\cite{IGEVStereo}, which is pre-trained on the KITTI-360 dataset in an unsupervised manner, to obtain predicted depth maps $D_{t}$. Next, we use the depth map, segmentation mask $S_{t}$, and the corresponding camera pose to generate the object point cloud $P_{t}^{n}$ through our designed depth to pseudo point cloud conversion, detailed in Algorithm~\ref{alg:1}. This process employs morphological operations with density-based filtering to remove unreliable outlines to retain reliable point clouds. The point clouds are further used to get the initial attributes of the 3D bounding boxes and the dynamic mask.  
 
\begin{algorithm}[!t]
    \small % 调整字体大小
    \caption{Depth to Pseudo Point Cloud Conversion}
    \label{alg:1}

    % 输入说明放在 algorithmic 外面当正文即可
    Inputs include stereo images $I_{t}^l$, $I_{t}^r$, segmentation masks $S_{t}$,
    camera intrinsics $K$, extrinsics $T_{t}$, and pseudo depth map $D_{t}$ at
    timestamp $t$. $B$ and $f$ are the baseline and focal length, respectively.
    For a scene, there are $T$ frames. $\eta_1$ and $\eta_2$ are thresholds.
    
    \vspace{1mm}
    \noindent\texttt{----------------------------------------}
    \vspace{-1mm}

    \begin{algorithmic}[1]
        \FOR{$t \in \{t_{0}, t_{1}, \dots, t_{T}\}$}
            \STATE \textbf{(1) Erode the segmentation mask to remove unreliable boundaries:}
            \STATE $S'_{t} \gets \mathrm{Erode}\left(S_{t}, \mathbf{K}_{k \times k} \right)$
            
            \STATE \textbf{(2) Minimal warping error selection:}
            \STATE $S_{t}'(u_x, u_y) \gets S'_{t}(u_x, u_y) \cdot M_{\text{warp}}$
            \STATE $M_{\text{warp}} \gets 
            \begin{cases} 
                1, & \text{if } \mathrm{warp}\!\left(I_l, I_r, \frac{Bf}{D_{t}}\right) \le \eta_{1}, \\
                0, & \text{otherwise}.
            \end{cases}$

            \STATE \textbf{(3) Project depth into 3D point cloud space:}
            \STATE $P_{t} \gets  \{\,T_{t} \cdot K^{-1} U \cdot D_{t} \mid S'_{t} > 0\,\}$

            \STATE \textbf{(4) Density-based filtering:}
            \STATE $P'_{t} \gets \{\,p_i \in P_t \mid \rho(p_i) \geq \eta_{2}\,\}$
        \ENDFOR
        \STATE \textbf{return} $P'$
    \end{algorithmic}
\end{algorithm}

% Velocity Computation.
\noindent \textbf{Velocity Initialization and Dynamic Mask Computation:} For the $n-th$ instance in the scene, the estimated object point cloud at time $t$ is denoted as $P_{t}^{n}$. We assume the velocity of this instance remains constant over time, allowing us to compute the velocity as:
\begin{align}
    v_{n}^{t} &= \frac{1}{\Delta{t}}(\frac{P_{t+\Delta{t}}^{n}}{M_{t+\Delta{t}}} - \frac{\text{ICP}(P_{t}^{n})}{M_{t}}),\\
    V_{n}^{\text{init}} &= \sum\frac{M_{t}\times v_{n}^{t}}{\sum_{t=0}^{T} M_{t}}, 
\end{align}
where \text{ICP} refers to the Iterative Closest Point~\cite{affineICP} algorithm, and $M_{t}$ denotes the number of point clouds associated with the current instance at time $t$. The dynamic mask $\widetilde{M}_{n}$ is then computed by selecting instances with velocities exceeding a predefined threshold of $0.03$ $m/s$. 

% Location Computation.
\noindent \textbf{Location Initialization:} The 3D bounding box location in the source frame $s$ is estimated by approximating its center as the average of the instances' point clouds. To mitigate the issue of sparse point clouds in the source frame, we move point clouds from adjacent frames $a$ using the estimated initial velocity $V_{\text{init}}$, thereby enhancing point cloud density for more accurate location estimation. This process is defined as:
\begin{equation}
L^{\text{init}}_{n} = \begin{cases} 
    \frac{1}{M_s}P_{n}^{s} & \text{if } M_{s} \geq 100, \\
    \frac{1}{M_a} {\left( P_{n}^{a} + V_{n}^{\text{init}} \cdot (t_s - t_{a}) \right)} & \text{if } M_{s} < 100
\end{cases}
\end{equation}
$M$ represents the number of point clouds in the source frame. 

% Orientation Computation.
\noindent \textbf{Orientation Initialization:} Orientation computation is divided into dynamic and static object orientations. For dynamic objects, $\theta_{n}^{\text{init}*}$, the orientation aligns with the velocity direction, and thus we initialize the dynamic orientation using the velocity direction. For static objects, $\theta_{n}^{\text{init}}$, we adopt the approach used in WeakM3D~\cite{WeakM3D}, where the overall orientation of the point cloud is utilized to approximate the orientation.

\subsection{Training of 3D Object Detectors}
\label{sec:method/training_of_3d_object_detectors}
After optimizing the 3D bounding boxes using the proposed multi-view 3D autolabeling approach, these boxes can be utilized as pseudo-labels to train 3D object detectors. To account for the quality of each pseudo-label during training, we introduce a simple yet effective confidence assignment method (Section~\ref{sec:method/training_of_3d_object_detectors/confidence_assignment}) and a confidence-based weighted loss for bounding box regression (Section~\ref{sec:method:loss_functions_for_monocular_3D_detectors}).

\begin{figure}[t]
    \setlength{\abovecaptionskip}{1mm}
    \begin{minipage}[b]{0.49\columnwidth}
        \centering
        \includegraphics[width=1.0\columnwidth]{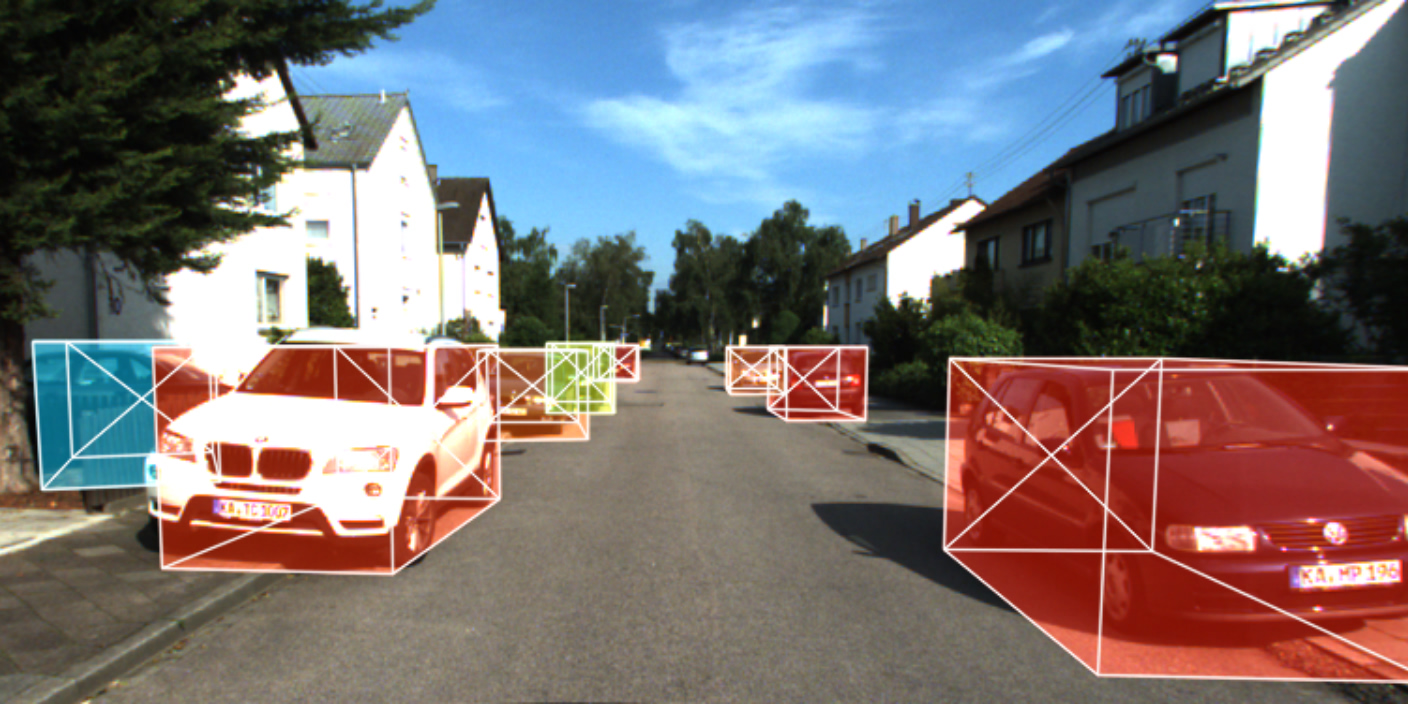}
        \subcaption{Static Scene}
    \end{minipage}
    \begin{minipage}[b]{0.49\columnwidth}
        \centering
        \includegraphics[width=1.0\columnwidth]{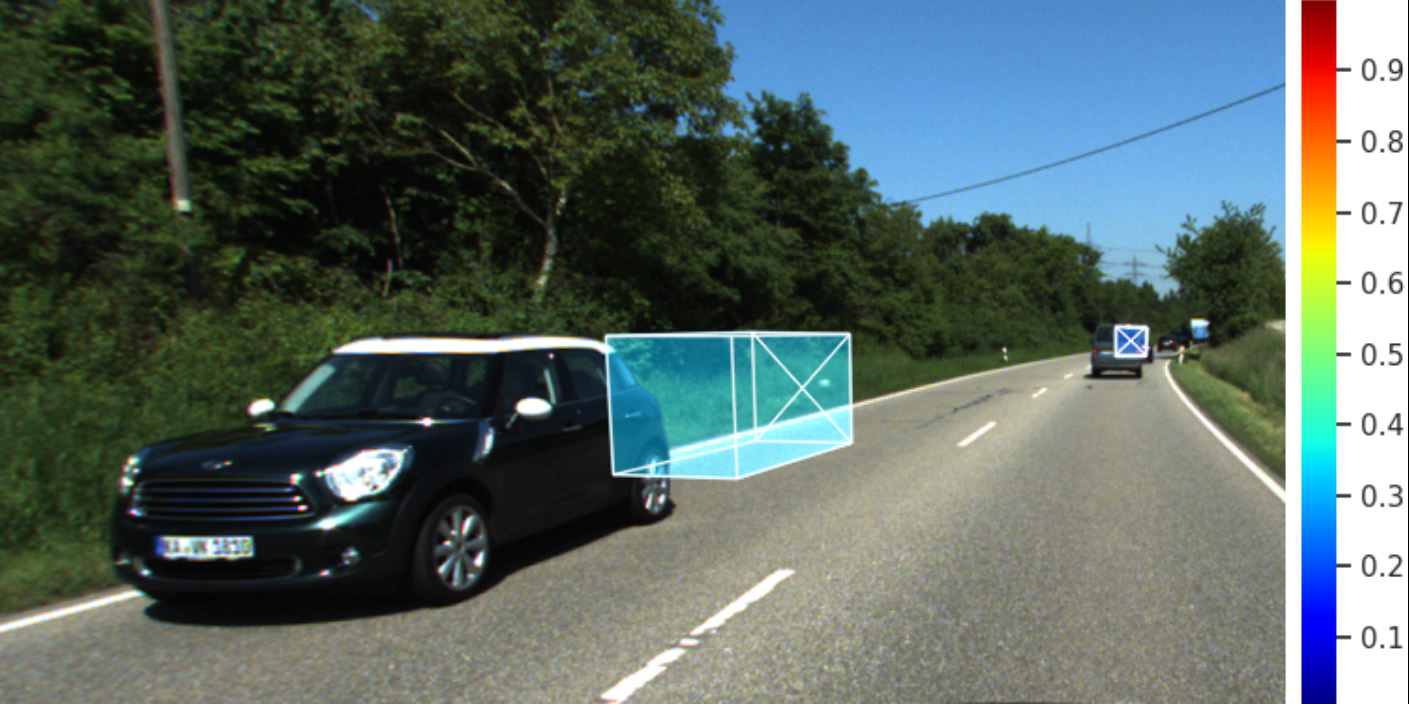}
        \subcaption{Dynamic Scene}
    \end{minipage}
    \caption{Comparison of the confidence scores. Lower confidence scores are observed for dynamic, occluded, or truncated objects, indicating their reduced influence on subsequent 3D object detector training.}
    \label{fig:method/training_of_3d_object_detectors/confidence_assignment}
\end{figure}

\subsubsection{Confidence Assignment}\label{sec:method/training_of_3d_object_detectors/confidence_assignment}
As shown in Figure~\ref{fig:method/training_of_3d_object_detectors/confidence_assignment}, the 3D bounding boxes optimized by the proposed autolabeling approach are not reliable for dynamic, occluded, or truncated objects. To address this issue, we propose a confidence assignment method based on the multi-view projection loss. 

First, for each target frame $t$, we identify a set of source frames $\mathcal{I}$ such that all instances in the target frame are visible from every frame in the set. Next, we determine an optimal bipartite matching between the optimized 3D bounding boxes and ground-truth 2D bounding boxes using the Hungarian algorithm. The cost matrix $\bm{Q} \in \mathbb{R}^{N \times N}$ is defined as the pairwise IoUs between the projected and ground-truth 2D bounding boxes, averaged over all the source frames, as follows: 
\begin{align}
\label{eq:method/training_of_3d_object_detectors/confidence_assignment/cost_matrix}
    \bm{Q}_{nm} =  \frac{1}{|\mathcal{I}|} \sum_{i \in \mathcal{I}} 1 - \text{IoU}(^{*}\!\bm{B}^{\text{2D}}_{in}, \bm{B}^{\text{2D}}_{im}) \ ,
\end{align}
where $^{*}\!\bm{B}^{\text{2D}}_{in}$ denotes the $n$-th projected 2D bounding box in frame $i$, obtained by projecting the $n$-th optimized 3D bounding box $^{*}\!\bm{B}_{n}$ onto frame $i$, and $\bm{B}^{\text{2D}}_{im}$ represents the $m$-th ground-truth 2D bounding box in frame $i$. Once an optimal permutation matrix $\bm{P} \in \{0, 1\}^{N \times N}$ is obtained, the confidence scores $^{*}\!\bm{C} \in [0, 1]^{N}$ for the optimized 3D bounding boxes $^{*}\!\bm{B}$ are computed as follows:
\begin{align}
    \label{eq:method/training_of_3d_object_detectors/confidence_assignment/confidence_scores}
    ^{*}\!\bm{C} = \frac{1}{|\mathcal{I}|} \sum_{i \in \mathcal{I}} \text{IoU}(^{*}\!\bm{B}^{\text{2D}}_{i}, \bm{P} \bm{B}^{\text{2D}}_{i}) \ .
\end{align}

\subsection{Loss Functions}\label{sec:method:loss_functions}
In this section, we introduce the loss functions used in the multi-view 3D autolabeling phase and the confidence-based weighted loss for training monocular 3D detectors.

\subsubsection{Loss Functions for Multi-View 3D Autolabeling}\label{sec:method:loss_functions_for_multi_view_3D_autolabelsing}
We optimize the parameters of the target 3D bounding boxes $\hat{\bm{\Omega}} = \{\hat{\bm{D}}, \hat{\bm{L}}, \hat{\bm{\Theta}}, \hat{\bm{V}}\}$ in the target frame along with the instance embeddings $\bm{Z}$ and the parameter $\bm{\psi}$ of the hypernetwork $\mathcal{H}(\cdot ; \bm{\psi})$. The final loss $\mathcal{L}$ is formulated as a combination of the \textit{multi-view projection loss} $\mathcal{L}_{\text{proj}}$, \textit{multi-view silhouette loss} $\mathcal{L}_{\text{slh}}$, Eikonal regularization $\mathcal{L}_{\text{eik}}$ \cite{NeuS}, and the \textit{attribute initialization regularization loss} $\mathcal{L}_{\text{init}}$ as follows:
\begin{align}
\label{sec:method/multi_view_3d_auto_labeling/loss_functions/total_loss}
    \mathcal{L}(\hat{\bm{\Omega}}, \bm{Z}, \bm{\psi}) 
    & = \lambda_{\text{proj}} \mathcal{L}_{\text{proj}}(\hat{\bm{\Omega}}) + \lambda_{\text{slh}} \mathcal{L}_{\text{slh}}(\hat{\bm{\Omega}}, \bm{Z}, \bm{\psi}) \nonumber  \\
    & + \lambda_{\text{eik}} \mathcal{L}_{\text{eik}}(\hat{\bm{\Omega}}, \bm{Z}, \bm{\psi}) + \lambda_{\text{init}} \mathcal{L}_{\text{init}}(\hat{\bm{\Omega}}) \ ,  
\end{align}
where $\lambda_{\text{proj}}$, $\lambda_{\text{slh}}$, $\lambda_{\text{eik}}$, and $\lambda_{\text{init}}$ denote the loss weights. 

Although $\mathcal{L}_{\text{proj}}$ and $\mathcal{L}_{\text{slh}}$ use 2D supervision, defining their 3D-2D correspondence is non-trivial. To address this, we employ the Hungarian algorithm \cite{Hungarian} for optimal bipartite matching between optimized 3D and ground-truth 2D bounding boxes. Each pairwise matching cost is defined as the projection loss introduced in this section but is computed only for the target frame rather than all source frames. For simplicity, we assume that the ground-truth 2D bounding boxes and instance masks have already been reordered based on the optimal permutation.

\noindent \textbf{Multi-View Projection Loss: }
Introducing the residual distance field (RDF) enables modeling arbitrary surfaces within a 3D bounding box. However, this also introduces a challenge: the 3D bounding box can expand indefinitely without constraints. To address this, we constrain the 3D bounding box using the ground-truth 2D bounding box to ensure tight alignment. Specifically, we define the multi-view projection loss, $\mathcal{L}_{\text{proj}}$, as the average distance between the projected 2D bounding box and the ground-truth, as follows:
\begin{align}
\label{eq:method/multi_view_3d_auto_labeling/loss_functions/multi_view_projection_loss/multi_view_projection_loss}
    \mathcal{L}_{\text{proj}}(\hat{\bm{\Omega}}) 
    & = \alpha \sum_{i \in \mathcal{S}} \sum_{n=1}^{N} \|\hat{\bm{B}}^{\text{2D}}_{in} - \bm{B}^{\text{2D}}_{in}\|_{\rm{H}} \nonumber \\
    &-\beta \sum_{i \in \mathcal{S}} \sum_{n=1}^{N} \text{DIoU}(\hat{\bm{B}}^{\text{2D}}_{in}, \bm{B}^{\text{2D}}_{in}) \ , 
\end{align}
where $\|\cdot\|_{\rm{H}}$ denotes the Huber loss, $\text{DIoU}(\cdot, \cdot)$ denotes the Distance-IoU \cite{DIoU}, and $\alpha, \beta$ are balancing coefficients. For projected 2D bounding box computation, unlike \cite{liu2024vsrd}, which assumes all sequences share the same 3D bounding box, we use the optimized velocity $\hat{v}_{n}$ to obtain a deformable 2D bounding box with the relative time gap $\Delta t$, indicating the relative frame ID difference between the current source frame and the target frame, described as follows:
\begin{align}
\hat{\bm{V}}^{\text{2D}}_{in} \propto \hat{\bm{B}}_{n} + (\hat{v}_{n} \Delta t) \bm{E}_{i}^{T} \bm{K}_{i}^{T},
\end{align}
\noindent where the projected 2D bounding box $\hat{\bm{B}}^{\text{2D}}_{in}$ is defined as the rectangle with the minimal area enclosing the projected vertices $\hat{\bm{V}}^{\text{2D}}_{in}$.$\bm{E}_{i}$ and $\bm{K}_{i}$ denote the extrinsic and intrinsic matrices for frame $i$, respectively.

% \paragraph{Multi-View Silhouette Loss} 
% \label{sec:method/multi_view_3d_auto_labeling/loss_functions/multi_view_silhouette_loss}
\noindent \textbf{Multi-View Silhouette Loss: }
The multi-view silhouette loss is defined as the cross-entropy between the rendered and ground-truth instance masks. Since the spatial gap between the surfaces of each instance and the 3D bounding box is modeled by the residual distance field introduced in Section~\ref{sec:method/multi_view_3d_auto_labeling_old/residual_distance_field}, the 3D bounding box constrained to tightly fit the ground-truth 2D bounding box by the multi-view projection loss $\mathcal{L}_{\text{proj}}$ is further refined by the multi-view silhouette loss $\mathcal{L}_{\text{slh}}$:
\begin{align}
\label{eq:method/multi_view_3d_auto_labeling/loss_functions/multi_view_silhouette_loss/multi_view_silhouette_loss}
    \mathcal{L}_{\text{slh}}(\hat{\bm{\Omega}}, \bm{Z}, \bm{\psi}) = \sum_{i \in \mathcal{S}} \sum_{j=1}^{R_{i}} \text{CE}(\hat{\bm{S}}(\bm{o}_{i}, \bm{d}_{ij}), \bm{S}_{ij}) \ , 
\end{align}
where $\bm{o}_{i} \in \mathbb{R}^{3}$, $\bm{d}_{ij} \in \mathbb{R}^{3}$, and $\bm{S}_{ij} \in \{0, 1\}^{N}$ denote the camera position, ray direction, and ground-truth instance label at the $j$-th sampled pixel in frame $i$, respectively. $\hat{\bm{S}}(\bm{o}_{i}, \bm{d}_{ij})$ denotes the rendered instance label based on Eq.~\eqref{eq:method/multi_view_3d_auto_labeling/instance_aware_volumetric_silhouette_rendering/integration}. $\text{CE}(\cdot, \cdot)$ denotes the cross-entropy loss. $R_{i}$ denotes the number of rays sampled for frame $i$.

% \paragraph{Regularization Loss}
% \label{sec:method/multi_view_3d_auto_labeling/loss_functions/regularization_loss} 
\noindent \textbf{Regularization Loss: }
To enhance the stability of the training process, we adopt the Eikonal regularization loss that is utilized in NeuS~\cite{NeuS}, to regularize the SDF during the training of the residual SDF network: 
\begin{align}
    \mathcal{L}_{\text{eik}} = \frac{1}{nm} \sum_{k,i} \left( \|\nabla \sigma(\mathcal{G}(\bm{p} ;  \bm{\phi}_{n}))\|_2 - 1 \right)^2
\end{align}  

Moreover, optimizing both location and velocity introduces high degrees of freedom, causing instability in early training. To mitigate this, we use initial values from 3D attribute initialization pipeline to serve as a regularization term, lightly weighted to prevent excessive deviation, where the $\mathcal{L}_{init}=\|\hat{\mathcal{L}} -\mathcal{L}^{init} \| + \|\hat{\mathcal{V}} -\mathcal{V}^{init} \|$.

\subsubsection{Confidence-Based Weighted Loss}\label{sec:method:loss_functions_for_monocular_3D_detectors}

The 3D bounding boxes optimized by the proposed multi-view 3D autolabeling serve as pseudo labels for most 3D object detectors, which leverage 3D bounding boxes and semantic class labels as supervision. This process does not require modifications to the architectures, loss functions, or training procedures, except for the incorporation of confidence-based loss weighting. We integrate the confidence scores computed by Eq.~\eqref{eq:method/training_of_3d_object_detectors/confidence_assignment/confidence_scores} exclusively into the regression loss. Given a regression loss function $\mathcal{L}_{\text{box}}$, its confidence-based weighted version $\tilde{\mathcal{L}}_{\text{box}}$ is defined as follows:
\begin{align}
    \label{sec:method/training_of_3d_object_detectors/confidence_based_weighted_loss/confidence_based_weighted_loss}
    \tilde{\mathcal{L}}_{\text{box}}(\hat{\bm{B}}, ^{*}\!\bm{B}, ^{*}\!\bm{C}) = \sum_{m=1}^{M} \ ^{*}\!\bm{C}_{\pi(m)} \mathcal{L}_{\text{box}}(\hat{\bm{B}}_{m}, ^{*}\!\bm{B}_{\pi(m)}) ,
\end{align}
where $\hat{\bm{B}}$ denotes the predicted 3D bounding boxes, $^{*}\!\bm{B}$ represents the 3D bounding boxes optimized by the proposed autolabeling, and $^{*}\!\bm{C}$ denotes the corresponding confidence scores. $M$ is the number of positive anchors, and $\pi(\cdot)$ represents the label assigner that maps the index of an anchor to that of the matched ground truth.

% Experiments
\section{Experiments}
\label{sec:experiments}

% Datasets Descriptions
\subsection{Dataset}
\label{sec:experiments/datasets}
To evaluate the proposed VSRD++, we conduct experiments on the KITTI-360 dataset~\cite{KITTI-360} for the reason that KITTI-
360 is the only public dataset that provides consistent camera
poses, GT instance masks, and GT 3D boxes for evaluation. Initially, we adopt the same dataset split as in ~\cite{liu2024vsrd}, dividing the KITTI-360 dataset into training (43,855 images), validation (1,173 images), and test sets (2,531 images). This split is referred to as the \textit{VSRD2024 Split}. However, the training set in the \textit{VSRD2024 Split} is heavily skewed toward static objects, relying on prior knowledge of object distribution. To better reflect real-world data, we introduce a new split, \textit{Casual Split}, with a dynamic-to-static object ratio of 1:4. For fair comparison, the test set remains the same as in the \textit{VSRD2024 Split}. We follow the same evaluation protocol as the KITTI dataset \cite{KITTI}. However, since occlusion and truncation labels are not available for the KITTI-360 dataset, unlike the KITTI dataset, we consider only two difficulty levels, namely \textit{Easy} and \textit{Hard}, based on whether the height of each ground-truth 2D bounding box is greater than $40$ and $25$ pixels, respectively. We evaluate our method only on the \textit{Car} category, following prior works~\cite{WeakM3D,liu2024vsrd}.

% Ablation Studies Tables.
\begin{table*}[t]
    \caption{Ablation study on the KITTI-360 training set to verify the effectiveness of each component in our proposed multi-view 3D auto-labeling. We use the \textit{Casual Split}'s training split to evaluate the performance of the pseudo label by reporting the average precision (AP) for overall regions and the dynamic regions, respectively. The "RDF" denotes the Residual Distance Field, "Velocity" represents the Velocity-Based Deformable SDF Modeling, and "Initialization" refers to the Self-Supervised 3D Attribute Initialization for 3D bounding box initialization. \textcolor{red}{\textbf{Red}} means best performing, \textbf{Bold} means the second.}
    \label{tab:experiments/ablation_study/multi_view_3d_auto_labeling}
    \centering
     \renewcommand{\arraystretch}{1.2}
    \setlength{\tabcolsep}{1.3mm}
    \scalebox{1.0}{
    \begin{tabular}{ccccc|cccc|cccc}
    \hline
    \multicolumn{5}{c|}{\multirow{2}{*}{\textbf{Components}}} & \multicolumn{4}{c|}{\textbf{Overall}} & \multicolumn{4}{c}{\textbf{Dynamic}} \\ \cline{6-13} 
    \multicolumn{5}{c|}{} & \multicolumn{2}{c|}{\textbf{AP$_{\text{BEV}}$/AP$_{\text{3D}}$@0.3}} & \multicolumn{2}{c|}{\textbf{AP$_{\text{BEV}}$/AP$_{\text{3D}}$@0.5}} & \multicolumn{2}{c|}{\textbf{AP$_{\text{BEV}}$/AP$_{\text{3D}}$@0.3}} & \multicolumn{2}{c}{\textbf{AP$_{\text{BEV}}$/AP$_{\text{3D}}$@0.5}} \\ \hline
    \textbf{$\mathcal{L}_{\text{proj}}$} & \textbf{$\mathcal{L}_{\text{slh}}$} & \textbf{RDF} & \textbf{Velocity} & \textbf{Initialization} & \textbf{Easy} & \multicolumn{1}{c|}{\textbf{Hard}} & \textbf{Easy} & \textbf{Hard} & \textbf{Easy} & \multicolumn{1}{c|}{\textbf{Hard}} & \textbf{Easy} & \textbf{Hard} \\ \hline
    \checkmark &  &  &  &  & 61.58/60.96 & \multicolumn{1}{c|}{60.87/58.97} & 40.94/30.84 & 38.07/28.47 & 7.83/5.29 & \multicolumn{1}{c|}{6.36/4.62} & 1.92/0.29 & 1.85/0.28 \\
    \checkmark & \checkmark &  &   &  & 64.10/62.10 &  \multicolumn{1}{c|}{63.35/\textbf{60.42}} & 40.64/30.54 & 37.61/28.01  & 8.14/5.68  & \multicolumn{1}{c|}{7.19/4.84} & 2.08/1.51  & 1.82/0.51 \\
    \checkmark & \checkmark & \checkmark &  &  & \textbf{65.54}/62.54 & \multicolumn{1}{c|}{\textbf{64.58}/55.30} & \textbf{41.18}/\textbf{31.14} & \textbf{39.42/29.98} & 8.07/5.13 & \multicolumn{1}{c|}{6.45/4.86} & 1.47/0.70 & 1.39/0.65 \\
    \checkmark & \checkmark & \checkmark & \checkmark &  & 65.05/\textbf{63.22} & \multicolumn{1}{c|}{61.97/58.99} & 38.81/29.04 & 38.43/\textcolor{red}{\textbf{30.46}} & \textbf{27.64/20.58} & \multicolumn{1}{c|}{\textbf{25.82/18.24}} & \textbf{7.14/4.76} & \textbf{6.11/4.42} \\
    \checkmark & \checkmark & \checkmark & \checkmark & \checkmark & \textcolor{red}{\textbf{77.95/75.20}} & \multicolumn{1}{c|}{\textcolor{red}{\textbf{77.29/74.50}}} & \textcolor{red}{\textbf{48.02/34.68}} & \textcolor{red}{\textbf{42.14}}/28.89 & \textcolor{red}{\textbf{51.33/46.87}} & \multicolumn{1}{c|}{\textcolor{red}{\textbf{57.81/48.42}}} & \textcolor{red}{\textbf{24.61/14.04}} & \textcolor{red}{\textbf{24.87/16.85}} \\ \hline
    \end{tabular}
    
    }
    \vspace{-0mm}
\end{table*}

% Ablation Studies Figures
\begin{figure*}[!t]
    \centering
    \includegraphics[width=1.0\linewidth]{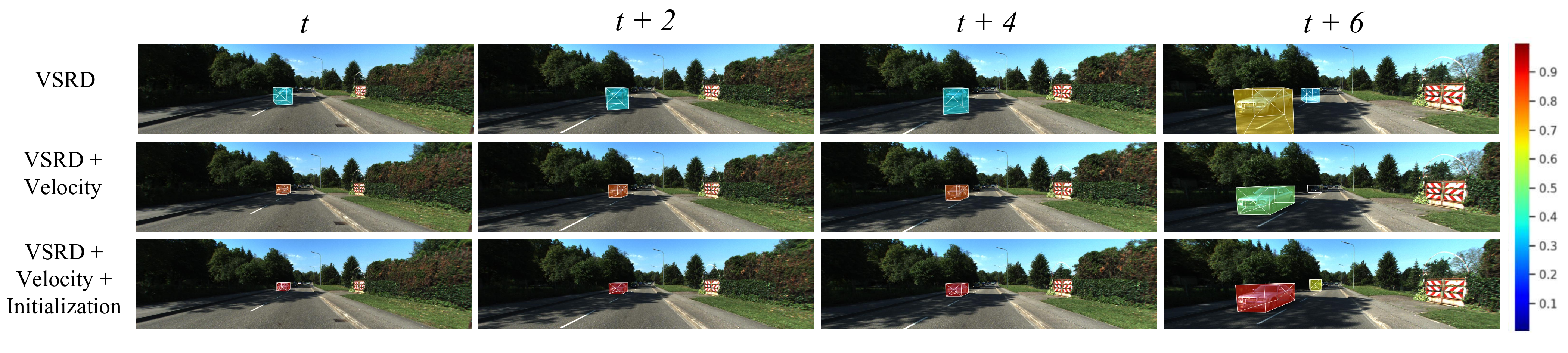}
  \caption{Visualization of pseudo labels from the ablation studies. The baseline(VSRD~\cite{liu2024vsrd}) corresponds to $\mathcal{L}_{\text{proj}} + \mathcal{L}_{\text{slh}} + \text{RDF}$, as proposed in ~\cite{liu2024vsrd}. Rows 2 and 3 illustrate the results after integrating \textit{Velocity-Based Deformable SDF Modeling} and further adding \textit{Self-Supervised 3D Attribute Initialization} for enhanced modeling of velocity objects. The visualizations show 3D bounding box changes over time, with colors indicating pseudo-label confidence: red for higher and blue for lower.
}
  \label{fig:experiments/evaluation_results/ablations}
\end{figure*}

\subsection{Implementation Details}
\label{sec:experiments/implementation_details}
\subsubsection{Multi-View 3D Autolabeling}
\label{sec:experiments/implementation_details/multi_view_3d_auto_labeling}
For both \textit{VSRD2024 Split} and the \textit{Casual Split}, we sample 16 source frames per target frame and select 1,000 rays per iteration based on ground truth instance masks (details in the supplementary material). Hierarchical volume sampling, as in NeRF~\cite{NeRF}, is used with 100 query points for both \textit{coarse} and \textit{fine} sampling. Unlike NeRF, all samples are drawn from a single scene SDF. Each instance embedding has a dimensionality of $D = 256$. The neural RDF $\mathcal{G}$ and hypernetwork $\mathcal{H}$ are MLPs with four hidden layers, containing 256 and 16 channels, respectively. We optimize the box parameters $\hat{\bm{\Omega}}$, instance embeddings $\bm{Z}$, and hypernetwork parameters $\bm{\psi}$ in $\mathcal{H}(\cdot; \bm{\psi})$ using the Adam optimizer~\cite{adam}. Learning rates decay exponentially from $1\text{e}{-2}$, $1\text{e}{-3}$, and $1\text{e}{-4}$ to $1\text{e}{-4}$, $1\text{e}{-5}$, and $1\text{e}{-6}$ over 3,000 iterations. Bounding box optimization is annealed over the first 1,000 iterations for dimensions, location, orientation, and velocity, after which RDF network training begins. Loss weights are set to $\alpha = 1.0$, $\beta = 0.1$, $\lambda_{\text{proj}} = 1.0$, $\lambda_{\text{slh}} = 1.0$, $\lambda_{\text{eik}} = 0.01$, and $\lambda_{\text{init}} = 0.1$. We optimize each scene for 3,000 iterations. To stabilize training, we adopt a warm-up schedule: during the first 1,000 iterations, we freeze the residual distance field (RDF) and optimize only the velocity and bounding-box parameters to obtain a good initialization of object locations and attributes; in the remaining 2,000 iterations, we jointly optimize all parameters, including the RDF network, to better model the scene geometry.

\subsubsection{Monocular 3D Object Detection}
\label{sec:experiments/implementation_details/monocular_3d_object_detection}
To thoroughly demonstrate the effectiveness of the pseudo labels generated during the multi-view autolabeling phase, we applied our generated pseudo 3D bounding boxes as pseudo labels to two widely adopted monocular 3D detectors: the CNN-based MonoFlex~\cite{MonoFlex} and the Transformer-based MonoDETR~\cite{MonoDETR}. For a comprehensive evaluation, we compared our method with the two-stage autolabeling approach AutoLabels~\cite{Autolabels}, applying the same training configurations for both MonoFlex and MonoDETR. To enhance compatibility, we adapted the original loss functions in MonoFlex and MonoDETR to the confidence-incorporated format, as detailed in Section~\ref{sec:method:loss_functions_for_monocular_3D_detectors}.  Additionally, we evaluate our approach against WeakM3D~\cite{WeakM3D}, a LiDAR-based weakly supervised 3D object detector, to evaluate its generalizability in a different 3D detection paradigm. To adapt WeakM3D to the KITTI-360 dataset, we utilized ground-truth 2D bounding boxes and instance masks for two key purposes: (1) applying RoIAlign~\cite{MaskRCNN} to extract object-specific features and (2) generating synthetic LiDAR points on object surfaces during training. These adaptations ensured compatibility with the KITTI-360 dataset while preserving the original framework's integrity. During inference, we employed Cascade Mask R-CNN~\cite{CascadeMaskRCNN} with InternImage-XL~\cite{InternImage} as an off-the-shelf 2D detector to generate high-quality 2D bounding boxes, which were subsequently used for RoIAlign operations in WeakM3D.

\subsection{Ablation Study}
\label{sec:experiments/ablation_study}

% Confidence Assignment Tables
\begin{table}[t]
    \caption{Ablation study on the KITTI-360 test set to verify the effectiveness of the confidence scores against monocular 3D object detection. MonoFlex~\cite{MonoFlex} is used as a monocular 3D object detector.}
    \label{tab:experiments/ablation_study/confidence_assignment}
    \centering
    \scalebox{1.0}{
        \begin{tabular}{ccccc}
            \toprule
            & \multicolumn{2}{c}{AP$_{\text{BEV}}$/AP$_{\text{3D}}$@0.3} & \multicolumn{2}{c}{AP$_{\text{BEV}}$/AP$_{\text{3D}}$@0.5} \\
            \cmidrule(lr){2-3} \cmidrule(lr){4-5}
            Conf. & Easy  & Hard  & Easy  & Hard  \\
            \midrule
                   & 51.19/43.14 & 44.20/40.44 & 24.60/17.37 & 22.37/16.25 \\
            \checkmark & \textbf{62.41}/\textbf{54.27} & \textbf{54.14}/\textbf{49.91} & \textbf{31.21}/\textbf{20.69} & \textbf{25.67}/\textbf{17.16} \\
            \bottomrule
        \end{tabular}
    }
    \vspace{-2mm}
\end{table}

% Compared with Others Autolabels Phase Tables
\begin{table*}[t]
    \caption{Evaluation results of our proposed multi-view 3D auto-labeling method, VSRD++, on the KITTI-360 training set. We compare its performance with the LiDAR-based monocular 3D auto-labeling method Autolabels~\cite{Autolabels} and the previous method VSRD~\cite{liu2024vsrd}. $^{*}$Results for Autolabels are reproduced using the official code. The comparison is conducted on both the \textit{VSRD2024 Split} and the \textit{Casual Split}. \textbf{Bold} means the best performance.}
    \label{tab:experiments/evaluation_results/multi_view_3d_auto_labeling}
    \centering
    \renewcommand{\footnoterule}{\empty}
    \renewcommand{\arraystretch}{1.5}
    \renewcommand{\thefootnote}{\fnsymbol{footnote}}
    \renewcommand{\thempfootnote}{\fnsymbol{mpfootnote}}
   \setlength{\tabcolsep}{1.8mm}
    \begin{tabular}{c|cc|cccc|cccc}
    \hline
    \multirow{3}{*}{Method} & \multicolumn{2}{c|}{\multirow{2}{*}{\begin{tabular}[c]{@{}c@{}}Weakly\\ Supervision\end{tabular}}} & \multicolumn{4}{c|}{\textit{VSRD2024 Split}} & \multicolumn{4}{c}{\textit{Casual Split}} \\ \cline{4-11} 
     & \multicolumn{2}{c|}{} & \multicolumn{2}{c|}{AP$_{\text{BEV}}$/AP$_{\text{3D}}$@0.3} & \multicolumn{2}{c|}{AP$_{\text{BEV}}$/AP$_{\text{3D}}$@0.5} & \multicolumn{2}{c|}{AP$_{\text{BEV}}$/AP$_{\text{3D}}$@0.3} & \multicolumn{2}{c}{AP$_{\text{BEV}}$/AP$_{\text{3D}}$@0.5} \\ \cline{2-11} 
     & LiDAR & Masks & Easy & \multicolumn{1}{c|}{Hard} & Easy & Hard & Easy & \multicolumn{1}{c|}{Hard} & Easy & Hard \\ \hline
    Autolabels*~\cite{Autolabels} & \checkmark & \checkmark  & 68.68/17.22 & \multicolumn{1}{c|}{66.65/11.88} & \textbf{52.18}/5.69 & 46.10/3.05 & 68.91/13.22 & \multicolumn{1}{c|}{67.20/11.60} & \textbf{51.86}/5.51 & 42.06/3.00 \\ \hline
    VSRD~\cite{liu2024vsrd} &  & \checkmark  & 75.03/68.53 & \multicolumn{1}{c|}{72.11/65.54} & 47.12/35.25 & 43.91/32.64 & 65.53/62.54 & \multicolumn{1}{c|}{64.58/55.30} & 41.17/31.44 & 39.42/29.98 \\
    VSRD++(Ours) &  & \checkmark & \textbf{87.20/77.10} & \multicolumn{1}{c|}{\textbf{78.60/76.19}} & 51.25/\textbf{36.12} & \textbf{48.83/34.23} & \textbf{77.95/75.20} & \multicolumn{1}{c|}{\textbf{77.29/73.50}} & 49.18/\textbf{34.58} & \textbf{42.15/28.89} \\ \hline
    \end{tabular}

\end{table*}

\begin{figure}[!t]
  \begin{minipage}[b]{1.0\linewidth}
    \centering
    \includegraphics[width=1.0\linewidth]{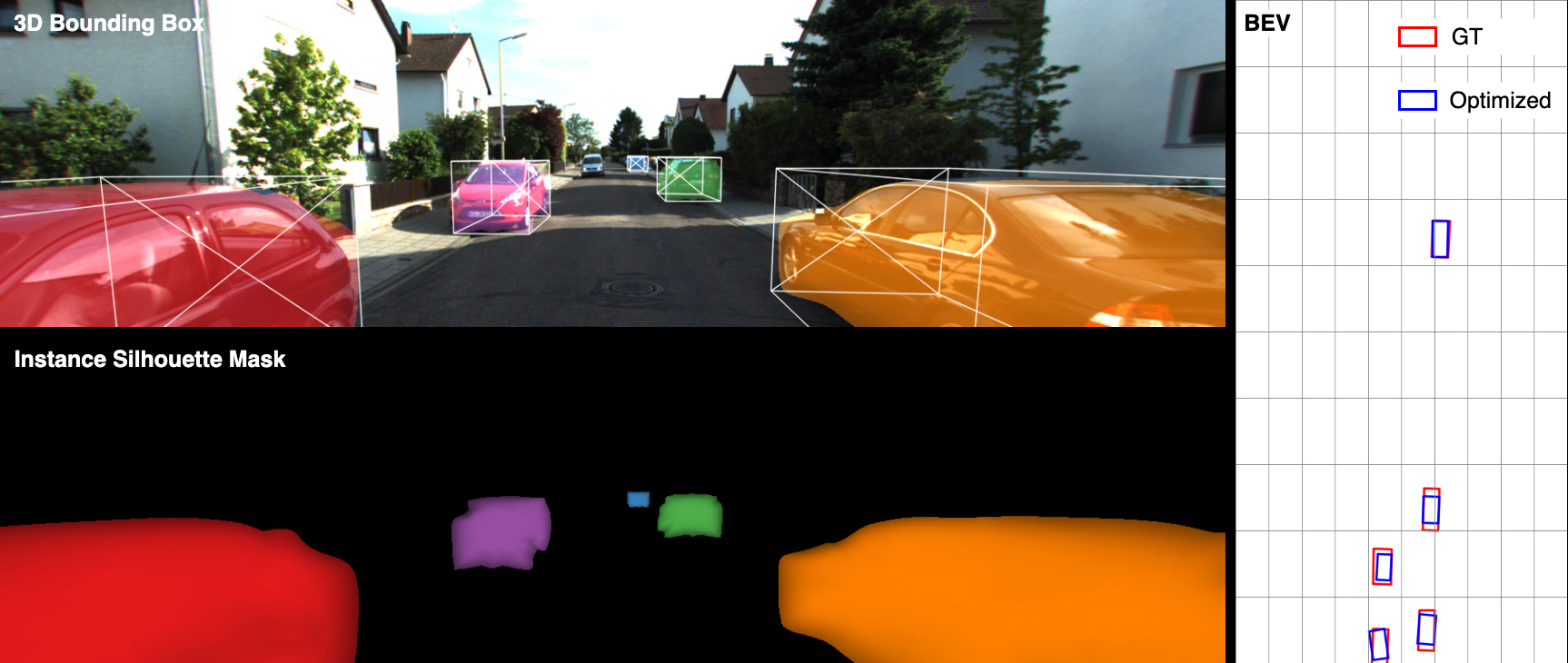}
  \end{minipage}
  \caption{Optimized 3D bounding boxes (1st row) and rendered instance masks (2nd row), with unique colors per instance and pixel colors as weighted sums of soft instance labels}
  \label{fig:experiments/evaluation_results/multi_view_3d_auto_labeling}
\end{figure}

\begin{table}[!t]
    \caption{Runtime and GPU memory usage of the multi-view autolabeling stage in VSRD++ for a typical KITTI-360 scene containing 6 car instances.}
    \label{tab:experiments/evaluation_results/autolabels/gpu_and_time}
    \centering
    \setlength{\tabcolsep}{1.1mm}
    \scalebox{0.9}{
        \begin{tabular}{c|c|c|c}
        \hline
        \textbf{Stage} & \textbf{Training Iterations} & 
        \textbf{\begin{tabular}[c]{@{}c@{}}Time Per Scene\\ (typical 6 cars)\end{tabular}} & 
        \textbf{GPU Memory} \\ \hline
        \begin{tabular}[c]{@{}c@{}}Box \& Velocity \\ Warm-up\end{tabular} & \textless{}1000 & 2 min 40s & 5.24GB \\ \hline
        \begin{tabular}[c]{@{}c@{}}Box, Velocity \\ \& RDF Refinement\end{tabular} & 1000$\sim$3000 & 16 min 31s & 11.7GB \\ \hline
        Total & 3000 & 18 min 11s & 11.7GB (max) \\ \hline
        \end{tabular}
    }
\end{table}

\subsubsection{Multi-View 3D Autolabeling}
\label{sec:experiments/ablation_study/multi_view_3d_auto_labeling}

We perform an ablation study to assess the effectiveness of each component in our proposed multi-view 3D autolabeling framework by evaluating the quality of the generated pseudo labels. To better simulate natural video sequences, we conduct experiments using the \textit{Casual Split}, where dynamic objects constitute 25\% of the dataset. The training set consists of 12,508 images, including 3,127 frames containing dynamic objects. Performance is measured by reporting the Average Precision (AP) in both BEV and 3D views across two evaluation settings: overall regions and dynamic regions.

As shown in Tab.~\ref{tab:experiments/ablation_study/multi_view_3d_auto_labeling}, the multi-view projection loss $\mathcal{L}_{\text{proj}}$ serves as a solid baseline on its own. The addition of the multi-view silhouette loss $\mathcal{L}_{\text{slh}}$ further enhances the quality of the pseudo labels to some extent. However, the spatial gap between the surfaces of each instance and the 3D bounding box constrains further improvements. To address this limitation, the residual distance field (RDF) significantly improves the quality of the pseudo labels. While this enhancement is evident in the \textit{Overall} regions, the average precision (AP) in dynamic regions remains relatively low. 

As illustrated in the first row of Figure~\ref{fig:experiments/evaluation_results/ablations}, the conventional SDF-based volume rendering (VSRD) struggles to represent moving vehicles due to inconsistencies in bounding box locations over time. By incorporating the velocity into 3D bounding box modeling, we observe a substantial performance boost, particularly for dynamic objects, with an increase in $\text{AP}_{\text{BEV}}@0.3$ from 8.07 to 27.64. Furthermore, incorporating the 3D attribute initialization pipeline allows us to fully exploit the potential of deformable rendering, achieving a significant improvement in dynamic object detection performance and raising the score to 51.23. 

The visualization results in Figure~\ref{fig:experiments/evaluation_results/ablations} demonstrate that by applying the velocity-based time-varying SDF modeling together with the 3D attribute initialization, we can generate high-confidence, continuous 3D bounding boxes that accurately track dynamic moving instances. Overall, the systematic improvements introduced by each component confirm their effectiveness, culminating in a highly precise autolabeling system for weakly supervised 3D object detection.

\subsubsection{Confidence Assignment}
\label{sec:experiments/ablation_study/confidence_assignment}

Due to the unreliability of the 3D bounding boxes optimized by the proposed autolabeling for dynamic, occluded, or truncated objects, we conducted an ablation study to demonstrate the effectiveness of our proposed confidence assignment. Tab.~\ref{tab:experiments/ablation_study/confidence_assignment} highlights substantial enhancement in detection performance when using the confidence-incorporated pseudo labels compared with the baseline, demonstrating the effectiveness of the proposed confidence assignment. 

\subsubsection{Computational Cost}
We further report the computational cost of the proposed multi-view autolabeling in
Table~\ref{tab:experiments/evaluation_results/autolabels/gpu_and_time} using a NVIDIA RTX 4090 GPU. For a typical KITTI-360 scene with six car instances, the warming-up stage (without RDF) terminates within fewer than 1000 iterations and requires approximately 2 minutes 40 seconds, with a peak memory footprint of 5.24 GB.
The subsequent stage runs for 1000–3000 iterations with an RDF hypernetwork, taking about 16~min~31~s and 11.7~GB of GPU memory. Overall, multi-view autolabeling of one such scene completes in roughly 18~min with a
maximum memory usage of 11.7~GB. This autolabeling procedure is executed offline, and its runtime grows approximately linearly with the number of object instances.

% Compared with Others in the Monocular 3D Detections Phase on the VSRD24 Splits
\begin{table*}[!t]
    \caption{Evaluation results of the stage-2: monocular 3D object detection on the KITTI-360 test set. All models are trained on \textit{VSRD24 Split}. $^{*}$Reproduced with the official code. $^{\dagger}$CAD models are used as extra data. $^{\ddagger}$$M(D)$ indicates that detection model $D$ is employed for model-agnostic method $M$. We split the methods into CNN-based and transformer-based. For each category, \textcolor{red}{\textbf{Red}} is best and \textbf{Bold} is second. (Except for the fully supervised manner colored in \textcolor{gray}{gray}.)}
    \label{tab:experiments/evaluation_results/monocular_3d_object_detection/comparsion_on_vsrd_split}
    \centering
    \renewcommand{\footnoterule}{\empty}
    \renewcommand{\thefootnote}{\fnsymbol{footnote}}
    \renewcommand{\thempfootnote}{\fnsymbol{mpfootnote}}
    \setlength{\tabcolsep}{1.8mm}
    \renewcommand{\arraystretch}{1.5}
    \scalebox{0.85}{
    \begin{tabular}{cc|ccc|c|cc|cc}
    \hline
    \multicolumn{2}{c|}{\multirow{2}{*}{Method}} 
        & \multicolumn{3}{c|}{Weakly Supervision} 
        & \multicolumn{1}{c|}{Full Supervision} 
        & \multicolumn{2}{c|}{AP$_{\text{BEV}}$/AP$_{\text{3D}}$@0.3} 
        & \multicolumn{2}{c}{AP$_{\text{BEV}}$/AP$_{\text{3D}}$@0.5} \\ \cline{3-10}
    \multicolumn{2}{c|}{} 
        & LiDAR & Mask & Pseudo Labels 
        & 3D Boxes 
        & Easy & Hard & Easy & Hard \\ \hline

    \multicolumn{1}{c|}{\multirow{5}{*}{\begin{tabular}[c]{@{}c@{}}CNN-Based\\ Method\end{tabular}}} 
        & WeakM3D~\cite{WeakM3D} 
        & \checkmark & \checkmark & 
        & 
        & 42.36/30.10 & 37.91/27.07 & 8.17/2.23 & 7.18/1.82 \\ \cline{2-10}
    \multicolumn{1}{c|}{} 
        & MonoFlex(Autolabels)~\cite{MonoFlex} 
        & \checkmark & \checkmark & Autolabels$^{\dagger}$~\cite{Autolabels} 
        & 
        & 45.77/31.57 & 35.48/26.93 & 27.87/9.10 & \textcolor{red}{\textbf{25.19}}/5.90 \\
    \multicolumn{1}{c|}{} 
        & MonoFlex(VSRD)~\cite{MonoFlex} 
        &  & \checkmark & VSRD~\cite{liu2024vsrd} 
        & 
        & \textbf{53.82/46.70} & \textbf{43.43/41.38} & \textbf{28.30/19.91} & 24.47/\textbf{16.34} \\
    \multicolumn{1}{c|}{} 
        & MonoFlex(VSRD++)~\cite{MonoFlex} 
        &  & \checkmark & \textbf{VSRD++(Ours)} 
        & 
        & \textcolor{red}{\textbf{57.03/54.27}} & \textcolor{red}{\textbf{51.85/42.20}} & \textcolor{red}{\textbf{32.99/22.41}} & \textbf{25.11}/\textcolor{red}{\textbf{16.37}} \\ 
    \multicolumn{1}{c|}{} 
        & \textcolor{gray}{MonoFlex(GT)}~\cite{MonoFlex} 
        &  &  &  
        & \checkmark
        & 69.70/67.07 & 59.86/57.26 & 50.82/43.11 & 41.78/34.43 \\ \hline

    \multicolumn{1}{c|}{\multirow{3}{*}{\begin{tabular}[c]{@{}c@{}}Transformer-Based\\ Method\end{tabular}}} 
        & MonoDETR(Autolabels)~\cite{MonoDETR} 
        &  & \checkmark & Autolabels$^{\dagger}$~\cite{Autolabels} 
        & 
        & 34.02/28.80 & 28.13/22.95 & 23.19/14.09 & 20.34/10.93 \\
    \multicolumn{1}{c|}{} 
        & MonoDETR(VSRD)~\cite{MonoDETR} 
        &  & \checkmark & VSRD~\cite{liu2024vsrd} 
        & 
        & \textbf{58.40/50.86} & \textbf{50.61/43.45} & \textbf{29.07/21.77} & \textbf{22.83/16.46} \\
    \multicolumn{1}{c|}{} 
        & MonoDETR(VSRD++)~\cite{MonoDETR} 
        &  & \checkmark & \textbf{VSRD++(Ours)} 
        & 
        & \textcolor{red}{\textbf{60.83/51.41}} & \textcolor{red}{\textbf{55.66/47.27}} & \textcolor{red}{\textbf{36.85/28.52}} & \textcolor{red}{\textbf{29.46/21.46}} \\ 
    \multicolumn{1}{c|}{} 
        & \textcolor{gray}{MonoDETR(GT)}~\cite{MonoDETR} 
        &  &  &  
        & \checkmark
        & 63.07/60.49 & 54.04/50.03 & 47.21/41.01 & 36.05/30.38 \\ \hline

    \end{tabular}}
\end{table*}

% Compared with Others in the Monocular 3D Detections Phase on the Casual Splits
\begin{table*}[!t]
    \caption{Evaluation results of the stage-2: monocular 3D object detection on the KITTI-360 test set. All models are trained on \textit{Casual Split}. $^{*}$Reproduced with the official code. $^{\dagger}$CAD models are used as extra data. $^{\ddagger}$$M(D)$ indicates that detection model $D$ is employed for model-agnostic method $M$. We split the methods into CNN-based and transformer-based. For each category, \textcolor{red}{\textbf{Red}} is best and \textbf{Bold} is second. (Except for the fully supervised manner colored in \textcolor{gray}{gray}.)}
\label{tab:experiments/evaluation_results/monocular_3d_object_detection/comparsion_on_casual}
    \centering
    \renewcommand{\footnoterule}{\empty}
    \renewcommand{\thefootnote}{\fnsymbol{footnote}}
    \renewcommand{\thempfootnote}{\fnsymbol{mpfootnote}}
    \setlength{\tabcolsep}{1.8mm}
    \renewcommand{\arraystretch}{1.5}
    \scalebox{0.85}{
    \begin{tabular}{cc|ccc|c|cc|cc}
    \hline
    \multicolumn{2}{c|}{\multirow{2}{*}{Method}} 
        & \multicolumn{3}{c|}{Weakly Supervision} 
        & \multicolumn{1}{c|}{Full Supervision} 
        & \multicolumn{2}{c|}{AP$_{\text{BEV}}$/AP$_{\text{3D}}$@0.3} 
        & \multicolumn{2}{c}{AP$_{\text{BEV}}$/AP$_{\text{3D}}$@0.5} \\ \cline{3-10}
    \multicolumn{2}{c|}{} 
        & LiDAR & Mask & Pseudo Labels 
        & 3D Boxes 
        & Easy & Hard & Easy & Hard \\ \hline

    \multicolumn{1}{c|}{\multirow{4}{*}{\begin{tabular}[c]{@{}c@{}}CNN-Based\\ Method\end{tabular}}} 
        & WeakM3D~\cite{WeakM3D} 
        & \checkmark & \checkmark & 
        & 
        & 39.71/27.23 & 34.83/24.18 & 7.42/2.05  & 6.66/1.78 \\ \cline{2-10}
    \multicolumn{1}{c|}{} 
        & MonoFlex(Autolabels)~\cite{MonoFlex} 
        & \checkmark & \checkmark & Autolabels$^{\dagger}$~\cite{Autolabels} 
        & 
        & 47.86/34.75 & 43.94/27.70 & \textcolor{red}{\textbf{36.50}}/12.97 & \textcolor{red}{\textbf{32.38}}/11.42 \\
    \multicolumn{1}{c|}{} 
        & MonoFlex(VSRD)~\cite{MonoFlex} 
        &  & \checkmark & VSRD~\cite{liu2024vsrd} 
        & 
        & \textbf{51.35/43.17} & \textbf{44.11/40.60} & 25.73/\textbf{16.01} & 22.75/\textbf{13.88} \\
    \multicolumn{1}{c|}{} 
        & MonoFlex(VSRD++)~\cite{MonoFlex} 
        &  & \checkmark & \textbf{VSRD++(Ours)} 
        & 
        & \textcolor{red}{\textbf{62.41/54.27}} & \textcolor{red}{\textbf{54.14/49.91}} & \textbf{31.21}/\textcolor{red}{\textbf{20.69}} & \textbf{25.67}/\textcolor{red}{\textbf{17.16}} \\ 
    \multicolumn{1}{c|}{} 
        & \textcolor{gray}{MonoFlex(GT)}~\cite{MonoFlex} 
        &  &  & 
        & \checkmark
        & 71.92/69.21 & 62.33/59.80 & 52.29/43.87& 45.66/38.12 \\ 

    \hline

    \multicolumn{1}{c|}{\multirow{3}{*}{\begin{tabular}[c]{@{}c@{}}Transformer-Based\\ Method\end{tabular}}} 
        & MonoDETR(Autolabels)~\cite{MonoDETR} 
        &  & \checkmark & Autolabels$^{\dagger}$~\cite{Autolabels} 
        & 
        & 35.27/30.25 & 28.70/26.39 & \textbf{27.68}/16.75 & \textbf{21.67}/14.23 \\
    \multicolumn{1}{c|}{} 
        & MonoDETR(VSRD)~\cite{MonoDETR} 
        &  & \checkmark & VSRD~\cite{liu2024vsrd} 
        & 
        & \textbf{46.66/38.86} & \textbf{42.24/34.97} & 24.72/\textbf{18.43} & 19.62/\textbf{15.04} \\
    \multicolumn{1}{c|}{} 
        & MonoDETR(VSRD++)~\cite{MonoDETR} 
        &  & \checkmark & \textbf{VSRD++(Ours)} 
        & 
        & \textcolor{red}{\textbf{56.03/47.95}} & \textcolor{red}{\textbf{47.73/43.81}} & \textcolor{red}{\textbf{27.85/19.35}} & \textcolor{red}{\textbf{24.69/17.34}} \\ 
    \multicolumn{1}{c|}{} 
        & \textcolor{gray}{MonoDETR(GT)}~\cite{MonoDETR} 
        &  &  &  
        & \checkmark
        & 62.50/60.73 & 61.96/59.48 & 50.88/47.79& 48.41/39.28 \\ 
        
        \hline
    \end{tabular}}

\end{table*}

% Compared with Others in the Monocular 3D Detections Phase on the VSRD24 Splits Visualizations
\begin{figure*}[!htp]
    \centering
    \includegraphics[width=0.98\linewidth]{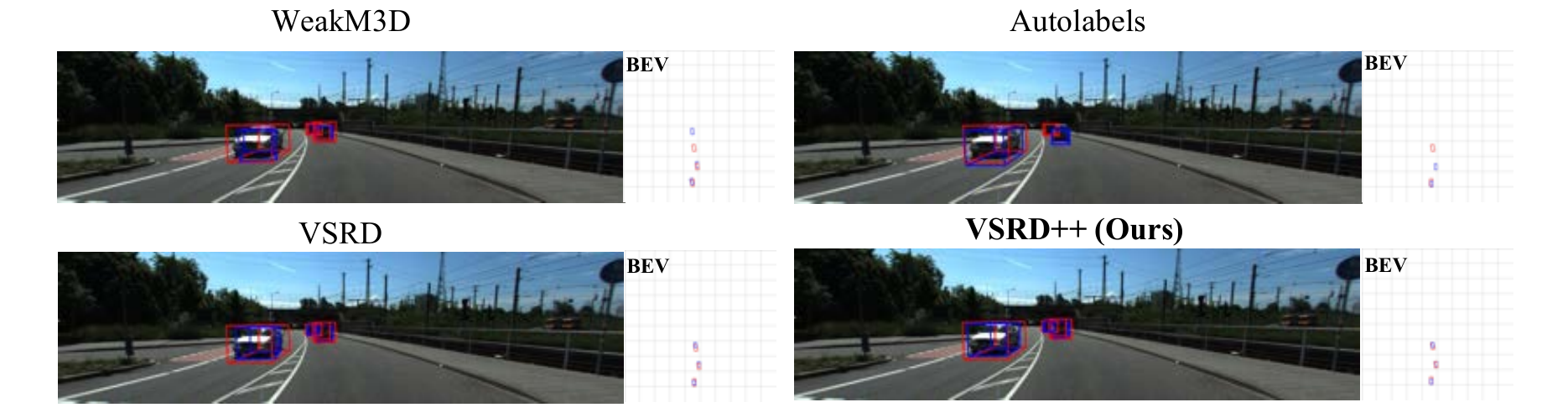}
  \caption{Visualization results of weakly supervised monocular 3D object detection results compared with Autolabels~\cite{Autolabels} and WeakM3D~\cite{WeakM3D} and VSRD~\cite{liu2024vsrd} with MonoFlex~\cite{MonoFlex} as the 3D detection backbone.
The ground truth and predicted bounding boxes are drawn in \textcolor{red}{red} and \textcolor{blue}{blue}, respectively.}
  \label{fig:experiments/evaluation_results/stage_2/compared_with_others}
\end{figure*}

% SDF Mesh
% SDF Decomposition
\begin{figure*}[!t]
    \centering
    \scalebox{0.85}{
    \includegraphics[width=\linewidth]{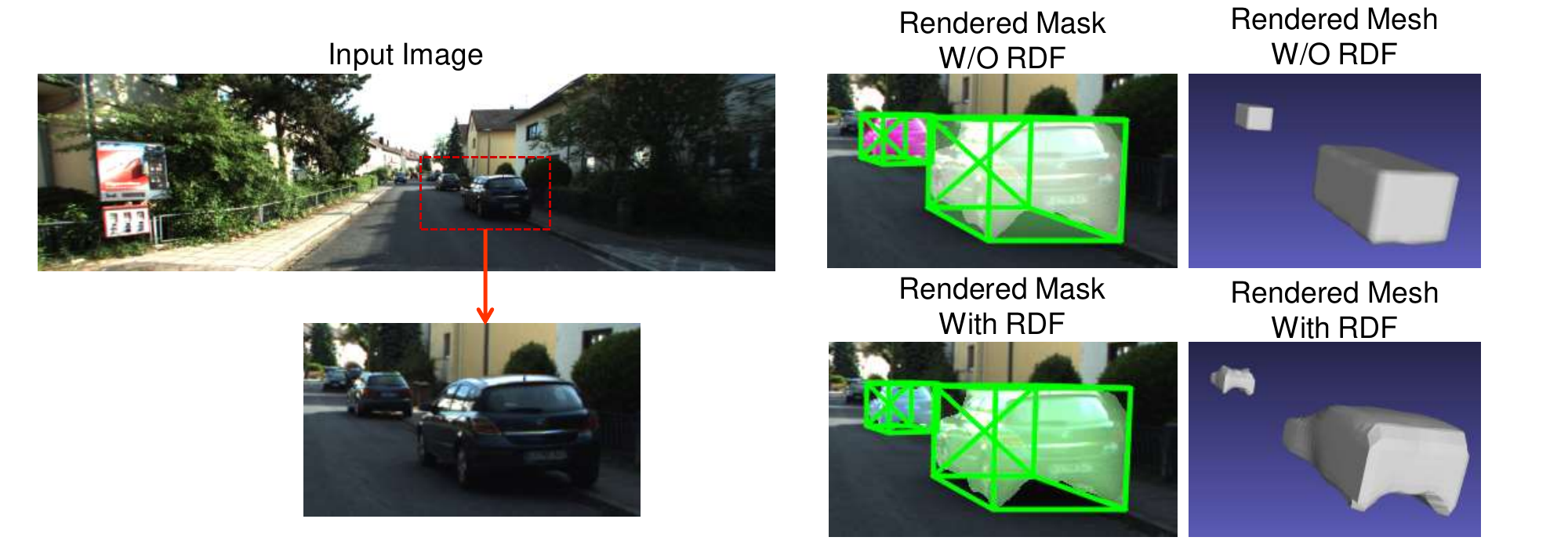}}
    \caption{Visualization of rendered masks and meshes using instance-aware volumetric rendering and Marching cubes~\cite{marching_cubes} with optimized SDF. The first row uses only the cuboid SDF, while the second includes the residual distance field (RDF). \textcolor{green}{Green} indicates optimized 3D bounding boxes by our multi-view 3D autolabeling.}
    \label{fig:sdf_mesh}
\end{figure*}

% Compared with Others in the Monocular 3D Detections Phase on the Semi-Supervised Settings
\begin{table}[t]
    \caption{Evaluation results of semi-supervised monocular 3D object detection on the KITTI validation set.}
    \label{tab:experiments/evaluation_results/monocular_3d_object_detection/semi_supervised_setting}
    \centering
    \setlength{\tabcolsep}{1.0mm}
    \scalebox{1.0}{
        \begin{tabular}{ccccc}
        \toprule
        & & \multicolumn{3}{c}{AP$_{\text{BEV}}$/AP$_{\text{3D}}$@0.7} \\
        \cmidrule(lr){3-5}        
        Method                           & Ratio & Easy                 & Moderate             & Hard                 \\
        \midrule
        MonoFlex \cite{MonoFlex}         & 1.00  & 33.49/22.94          & 23.83/16.05          & 20.27/13.29          \\ 
        \midrule
        \multirow{5}{*}{VSRD++ (MonoFlex)} & 0.00 & 3.73/1.31          & 2.07/0.52          & 1.46/0.36          \\
                                         & 0.25  & \textbf{35.61/25.19 }         & \textbf{25.91/17.42}          & \textbf{22.73/15.29 }         \\
                                         & 0.50  & \textbf{37.77/27.26} & \textbf{26.86/18.40} & \textbf{23.33/16.97} \\
                                         & 0.75  & \textbf{41.41/29.87} & \textbf{30.43/21.11} & \textbf{26.84/18.09} \\
                                         % & 1.00  & 40.66/30.80          & 30.21/23.05          & 25.53/19.73          \\ 
        \bottomrule
        \end{tabular}
    }
    \vspace{-4mm}
\end{table}

\subsection{Evaluation Results}
\label{sec:experiments/evaluation_results}
% Multi-View 3D Auto-Labeling
\subsubsection{Multi-View 3D Autolabeling}
\label{sec:experiments/evaluation_results/multi_view_3d_auto_labeling}
To further evaluate the quality of the generated pseudo labels, we compare our approach with two representative two-stage frameworks that consist of autolabeling followed by training 3D object detectors on the pseudo labels. Specifically, we compare against a degraded version of our method, \textit{VSRD}~\cite{liu2024vsrd}, and the LiDAR-based \textit{Autolabels}~\cite{Autolabels} to comprehensively assess the effectiveness of our pseudo-labeling strategy. Following the same experimental settings as the ablation studies, we conduct comparative experiments using two splits of the KITTI-360 dataset. The \textit{VSRD2024 Split} focuses on dynamic objects, while the \textit{Casual Split} better represents real-world autonomous driving scenarios with a considerable number of dynamic objects. As shown in Table~\ref{tab:experiments/evaluation_results/multi_view_3d_auto_labeling}, on the \textit{VSRD2024 Split}, our proposed VSRD++ significantly outperforms both \textit{Autolabels} and \textit{VSRD}, achieving improvements of 26.97\% and 16.22\%, respectively, in terms of $\text{AP}_{\text{BEV}}@0.3$. In the \textit{Casual Split}, where dynamic objects are more prevalent, the static modeling-based \textit{VSRD} exhibits inferior performance compared to \textit{Autolabels} in both $\text{AP}_{\text{BEV}}@0.3$ and $\text{AP}_{\text{BEV}}@0.5$. However, by incorporating velocity-based optimization and 3D attribute initialization \textit{VSRD++} achieves significant improvements in modeling dynamic objects, outperforming \textit{Autolabels} in the \textit{Casual Split} as well. The optimized 3D bounding boxes and rendered instance masks are visualized in Figure~\ref{fig:experiments/evaluation_results/multi_view_3d_auto_labeling}.

% Monocular 3D Object Detection.
\subsubsection{Monocular 3D Object Detection}
\label{sec:experiments/evaluation_results/monocular_3d_object_detection}
Although the proposed Multi-View 3D Autolabeling framework can generate high-quality pseudo labels, the optimization process typically requires 5–10 minutes to produce a reliable bounding box, which limits its practicality for real-time autonomous driving applications. To overcome this limitation, we further leverage the pseudo labels generated by the proposed autolabeling to serve as 3D supervision to investigate their applicability using the existing monocular 3D object detectors. We use the monocular 3D object detection for the reason that it is the most label-dependent and ill-conditioned
setting, thus providing the clearest testbed for evaluating the
benefit of our pseudo labels

% \paragraph{Weakly Supervised Setting} 
% \label{sec:experiments/evaluation_results/monocular_3d_object_detection/weakly_supervised_setting}
\noindent \textbf{Weakly Supervised Setting:} Table.~\ref{tab:experiments/evaluation_results/monocular_3d_object_detection/comparsion_on_vsrd_split} and Table.~\ref{tab:experiments/evaluation_results/monocular_3d_object_detection/comparsion_on_casual} shows the evaluation results of VSRD++ compared with the existing methods by integrate into different monocular 3D detection backbones on \textit{VSRD2024 Split} and \textit{Casual Split}, respectively. Our method demonstrates a significant superiority over WeakM3D \cite{WeakM3D} across all the metrics while eliminating the need for LiDAR points for 3D supervision. Moreover, the detector trained on the pseudo labels generated by the proposed autolabeling outperforms that trained on the pseudo labels generated by Autolabels \cite{Autolabels} on both static-dominated splits and the dynamic-dominated split by employing more sophisticated monocular 3D object detectors such as MonoFlex \cite{MonoFlex} and MonoDETR \cite{MonoDETR} further improves detection performance, demonstrating the broad versatility of our method, which is not limited to a specific detection model. As shown in Table~\ref{tab:experiments/evaluation_results/monocular_3d_object_detection/comparsion_on_vsrd_split}, in static-dominated splits, the 3D detectors trained with pseudo labels generated by the proposed \textit{VSRD++} outperform those trained with \textit{Autolabels} and \textit{WeakM3D}. However, the performance improvement over the original VSRD is relatively marginal, with AP$_\text{BEV}@0.3$ (Easy) increasing by 3.21 (5.96\%) and 2.43 (4.16\%) for MonoFlex and MonoDETR, respectively. In contrast, as shown in Table~\ref{tab:experiments/evaluation_results/monocular_3d_object_detection/comparsion_on_casual}, in dynamic-dominated splits, the detectors trained with VSRD++ pseudo labels achieve substantial improvements over VSRD, with AP$_\text{BEV}@0.3$ (Easy) increasing by 11.06 (21.5\%) and 9.37 (20.1\%) for MonoFlex and MonoDETR, respectively. These results highlight the enhanced robustness of VSRD++ in dynamic environments, demonstrating its effectiveness in addressing the challenges posed by object motion and temporal inconsistencies.  It is noteworthy that the detectors trained on the pseudo labels generated
by our method demonstrate competitive performance compared with those trained in a fully supervised manner as shown in Table~\ref{tab:experiments/evaluation_results/monocular_3d_object_detection/comparsion_on_casual} and Table~\ref{tab:experiments/evaluation_results/monocular_3d_object_detection/comparsion_on_vsrd_split}.

% \paragraph{Semi-Supervised Setting} 
% \label{sec:experiments/evaluation_results/monocular_3d_object_detection/semi_supervised_setting}
\noindent \textbf{Semi-Supervised Setting: }The essential advantage of our method is that it avoids costly 3D annotations, making more data available for training. Therefore, we investigate a realistic scenario where a detector pre-trained on a large amount of unlabeled data from a source domain is fine-tuned on a small amount of labeled data from a target domain. We select the KITTI-360 and KITTI datasets as source and target domains, respectively. Tab.~\ref{tab:experiments/evaluation_results/monocular_3d_object_detection/semi_supervised_setting} shows the performance of the detector pre-trained on the KITTI-360 dataset in a weakly supervised manner with the proposed autolabeling and then fine-tuned on a subset of the KITTI dataset in a supervised manner. The zero-shot performance is quite low due to the characteristic that monocular depth estimation is greatly affected by the differences in camera parameters, but the performance of the detector fine-tuned on only 25\% of the labeled data significantly outperforms that trained on the whole data from scratch, highlighting the broad applicability of our method.

\section{Limitations}
\label{limitations}

\noindent\textbf{Category coverage.}
In line with most existing weakly supervised monocular 3D detection methods such as WeakM3D and Autolabels, our experiments focus on the \textit{Car} category, which is the primary class in autonomous driving benchmarks.
Extending VSRD++ to pedestrians and cyclists is non-trivial: these objects are non-rigid and can change shape across the sequential views used in our multi-view setting, violating the near-rigid, single-SDF assumption in our volumetric modeling.
Moreover, KITTI-360 provides much sparser LiDAR annotations for these classes. These factors hinder stable multi-view SDF reconstruction and fair quantitative evaluation beyond cars.
Designing SDFs or alternative volumetric representations that better accommodate non-rigid, small objects is left for future work.

\begin{table}[!t]

\caption{Impact of segmentation quality and pseudo labels. We compare our framework using ground-truth (GT) instance masks and pseudo masks generated by GroundingSAM2.
For GT masks, we further simulate degraded segmentation by applying morphological erosion so that the foreground coverage with respect to the \emph{image area} is reduced by 1\%, 3\%, and 5\%. We report AP$_{\text{BEV}}$ and AP$_{3\text{D}}$ at IoU 0.3 on the Easy and Hard difficulty levels.}

\label{tab:experiments/evaluation_results/seg_ablations}
\centering
\setlength{\tabcolsep}{1.3mm}

\begin{tabular}{c|c|cc|cc}
\hline
\multirow{2}{*}{\textbf{Mask Source}} & \multirow{2}{*}{\textbf{Erosion Settings}} & \multicolumn{2}{c|}{\textbf{$\text{AP}_{\text{BEV}}@0.3$}} & \multicolumn{2}{c}{\textbf{$\text{AP}_{\text{3D}}@0.3$}} \\ \cline{3-6} 
 &  & \textbf{Easy} & \textbf{Hard} & \textbf{Easy} & \textbf{Hard} \\ \hline
\begin{tabular}[c]{@{}c@{}}GT Masks\\ (Original)\end{tabular} & No Erosion & \textbf{79.75} & \textbf{78.32} & \textbf{77.57} & \textbf{75.90} \\ \hline
\begin{tabular}[c]{@{}c@{}}Pseudo Masks\\ (GroundingSAM2~\cite{SAM2})\end{tabular} & No Erosion & 74.59 & 73.43 & 65.50 & 63.58 \\ \hline
\begin{tabular}[c]{@{}c@{}}GT Masks\\ (Original)\end{tabular} & \begin{tabular}[c]{@{}c@{}}1\% Area\\ Reduction\end{tabular} & 75.76 & 66.63 & 63.95 & 60.11 \\ \hline
\begin{tabular}[c]{@{}c@{}}GT Masks\\ (Original)\end{tabular} & \begin{tabular}[c]{@{}c@{}}3\% Area\\ Reduction\end{tabular} & 59.81 & 51.93 & 40.28 & 39.96 \\ \hline
\begin{tabular}[c]{@{}c@{}}GT Masks\\ (Original)\end{tabular} & \begin{tabular}[c]{@{}c@{}}5\% Area\\ Reduction\end{tabular} & 42.33 & 41.85 & 32.02 & 31.76 \\ \hline
\end{tabular}
\end{table}

\noindent\textbf{Dependence on 2D segmentation quality.}
Dependence on 2D segmentation quality. VSRD++ also relies on reasonably accurate 2D instance segmentation to provide silhouette-based self-supervision. As shown in Table~\ref{tab:experiments/evaluation_results/seg_ablations}, when we synthetically erode the ground-truth masks so that their foreground coverage w.r.t. the image area is reduced by 1\% of the total image size, the performance drops only moderately, while more aggressive 3\%–5\% erosion leads to a substantial degradation in both AP${\text{BEV}}$ and AP${3\text{D}}$. Pseudo masks generated by GroundingSAM2~\cite{GroundedSAM,SAM2} yield slightly lower but still competitive performance, indicating that recent high-performance segmentation foundation models provide sufficiently reliable masks on KITTI-360 for the Car class. However, the dependence on 2D segmentation quality is not completely removed. Future work may incorporate more robust multi-view consistency, joint refinement of 2D masks and 3D geometry, or additional weak cues such as depth, motion, or language priors to further reduce this sensitivity.

% Conclusibs
\section{Conclusion}
\label{sec:conclusion}

In this paper, we propose \textbf{VSRD++}, an enhanced weakly supervised framework that extends VSRD to dynamic environments, addressing its limitations in handling moving objects. By leveraging neural-field-based volumetric rendering with weak 2D supervision, VSRD++ eliminates the need for expensive 3D annotations while improving pseudo-label quality for both static and dynamic objects. To model dynamic scenes, we extend 3D bounding boxes with a velocity attribute, enabling time-varying SDF modeling, and propose a 3D attribute initialization pipeline that leverages pseudo-depth cues to enhance optimization stability. These advancements improve the robustness and accuracy of weakly supervised 3D object detection in complex scenes. Extensive experiments on the KITTI-360 dataset demonstrate that VSRD++ achieves superior performance compared to LiDAR-based Autolabels and the original VSRD, particularly in dynamic scenarios. By generalizing VSRD to broader applications, VSRD++ offers a scalable and practical solution for autonomous driving without relying on 3D ground truth. Future work will explore improvements in efficiency and broader applications across diverse real-world scenes.

\bibliography{bibtex/bib/IEEEabrv.bib,bibtex/bib/IEEEexample.bib}{}
\bibliographystyle{IEEEtran}

\vfill

\end{document}